\documentclass[twoside]{article}

\usepackage[accepted]{aistats2024}

\usepackage{natbib}
\usepackage{amsmath, amsfonts, amssymb, amsthm, bm, bbm, graphicx}
\usepackage{xcolor}
\usepackage[title]{appendix}
\usepackage{algorithm}
\usepackage{algpseudocode}
\usepackage{hyperref}

\allowdisplaybreaks

\usepackage{caption, subcaption}

\newtheorem{assumption}{Assumption}
\newtheorem*{assumption*}{Assumption}
\newtheorem{lemma}{Lemma}
\newtheorem*{lemma*}{Lemma}
\newtheorem{theorem}{Theorem}
\newtheorem*{theorem*}{Theorem}
\newtheorem{proposition}{Proposition}
\theoremstyle{remark}
\newtheorem{remark}{Remark}

\newcommand{\muTreated}{\mu_1\left(q \mid F_{\bm{x}_i}\right)}
\newcommand{\muControl}{\mu_0\left(q \mid F_{\bm{x}_i}\right)}
\newcommand{\muT}{\mu_t\left(q \mid F_{\bm{x}_i}\right)}
\newcommand{\hatmuT}{\hat{\mu}_t\left(q \mid F_{\bm{x}_i}\right)}
\newcommand{\sigmaT}{\sigma_t^2\left(q \mid F_{\bm{x}}\right)}

\begin{document}

\twocolumn[

\aistatstitle{Interpretable Causal Inference for Analyzing Wearable, Sensor, and Distributional Data}

\aistatsauthor{ Srikar Katta \And Harsh Parikh \And  Cynthia Rudin \And Alexander Volfovsky }

\aistatsaddress{ Duke University \And  
Johns Hopkins University \And Duke University \And Duke University } ]

\begin{abstract}
  Many modern causal questions ask how treatments affect complex outcomes that are measured using wearable devices and sensors. Current analysis approaches require summarizing these data into scalar statistics (e.g., the mean), but these summaries can be misleading. For example, disparate distributions can have the same means, variances, and other statistics. Researchers can overcome the loss of information by instead representing the data as distributions. We develop an interpretable method for distributional data analysis that ensures trustworthy and robust decision-making: Analyzing Distributional Data via Matching After Learning to Stretch (ADD MALTS). We (i) provide analytical guarantees of the correctness of our estimation strategy, (ii) demonstrate via simulation that ADD MALTS outperforms other distributional data analysis methods at estimating treatment effects, and (iii) illustrate ADD MALTS' ability to verify whether there is enough cohesion between treatment and control units within subpopulations to trustworthily estimate treatment effects. We demonstrate ADD MALTS' utility by studying the effectiveness of continuous glucose monitors in mitigating diabetes risks.
\end{abstract}

\section{\MakeUppercase{Introduction}} \label{sec:intro}
Diabetes -- a disease limiting glucose regulation in the bloodstream -- affects millions worldwide. According to the World Health Organization, Diabetes caused 2 million deaths in 2019 and is a leading cause of blindness, kidney failure, and heart attacks \citep{WHO_2023}. Continuous glucose monitors (CGMs), which are wearable devices that automatically track patients' blood glucose concentrations over time, offer a new avenue for diabetes care. CGMs allow researchers and clinicians to screen patients, propose treatments, and manage diets \citep{matabuena2021glucodensities, janine2008use, hall2018glucotypes, lu2021time}.

\begin{figure}[t]
    \centering
    \includegraphics[width=0.45\textwidth]{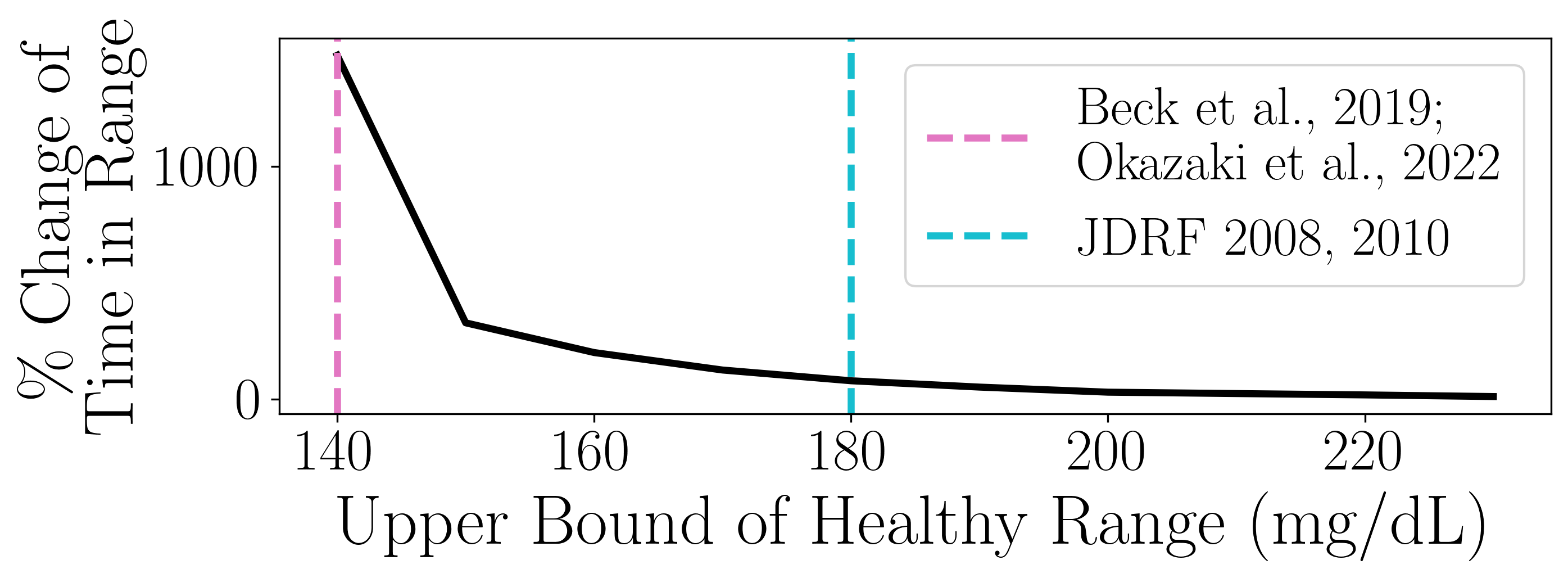}
    \caption{For patients older than 55 years, we measure the effectiveness of CGMs as the percent change of time in healthy range. The plot shows how changing the healthy range's upper bound (x-axis) affects the treatment effect (y-axis). 70 is the lower bound.}
    \label{fig:bounds_effect_lower70}
\end{figure}
While CGMs show much promise for diabetes care, the standard approaches for summarizing CGM data can lead to very misleading insights. To demonstrate these issues, we reanalyze CGM data from a study conducted by the Juvenile Diabetes Research Foundation (JDRF). JDRF ran a randomized experiment to investigate the effectiveness of CGMs in mitigating the risks of Diabetes using a cohort of 450 patients with type 1 diabetes. 
Each CGM's continuous stream of data was summarized by measuring how often a patient's blood glucose concentration was within a healthy range of 70-180 mg/dL. The treatment effect was then calculated by comparing pre-and-post ``time in range'' (TIR) between treated and control patients. While JDRF researchers used 70-180 mg/dL as a healthy range, slightly changing the healthy range to 70-140 mg/dL -- as used by \citet{okazaki2022association, beck2019relationships} -- completely changes the results. As shown in Figure \ref{fig:bounds_effect_lower70}, for patients older than 55 years old, \textbf{using the 70-140 mg/dL range would suggest that CGMs are 1300 percentage points more effective than if the healthy range was 70-180 mg/dL.} This case study highlights how summarizing complex CGM data using scalar statistics can lead to misleading insights, which may be detrimental to patient care.

To overcome the issues of scalar metrics, several researchers have recommended representing data from wearable devices as \textit{distributions}
\citep{matabuena2021glucodensities, ghosal2023shape, ghodrati2022distribution}. Rather than asking, ``how often is a patient's glucose concentration in a \textit{pre-described} healthy range,'' the distributional representation answers the question, ``how often is a patient's glucose concentration at \textit{any particular level for all possible levels}.'' \citet{matabuena2021glucodensities} demonstrates that the distributional representation of glucose concentrations is much richer than TIR, is clinically useful, and is highly correlated with other clinical biomarkers. 

Taking inspiration from the optimal transport literature \citep{vallender1974calculation}, \citet{lin2023causal} propose estimands and estimators for conducting causal inference with distributional outcomes, enabling us to derive rich insights from CGM data. However, these approaches rely on strong and often untestable assumptions. For example, the positivity assumption requires enough cohesion between treated and control units across subregions of the covariate space. When such assumptions fail, these techniques can yield misleading insights. For proper diabetes care and management, researchers require techniques that can help validate whether the strict assumptions in causal inference can hold. To this aim, we develop an end-to-end interpretable causal approach for analyzing distributional data: Analyzing Distributional Data via Matching After Learning to Stretch (ADD MALTS). 

\paragraph{Contributions} We prove that ADD MALTS can consistently estimate treatment effects with complex, distributional data. Via simulation, we demonstrate that ADD MALTS can more accurately estimate conditional average treatment effects than competing methods; we also show how ADD MALTS adds trustworthiness in the causal pipeline by validating whether treated and control units are comparable in subregions of the covariate space. Finally, we re-analyze data studying the effectiveness of CGMs in managing health risks in patients with type 1 diabetes, finding important insights about the data and CGMs.\footnote{All code to replicate simulation and real data analysis results are available here: \href{https://github.com/almost-matching-exactly/addmalts}{https://github.com/almost-matching-exactly/addmalts}}

\section{\MakeUppercase{Background}} \label{sec:background}
In this section, we introduce background concepts that are necessary for ADD MALTS. We first discuss the Wasserstein distance, which measures distances between distributions. Next, we discuss how we can ``average'' distributions (referred to as the barycenter). Finally, we connect the concepts from the Wasserstein space to ideas from traditional causal inference.

\paragraph{Wasserstein Space} Our work relies on the Wasserstein metric space for measuring distances between distributions \citep{vallender1974calculation}. The 2-Wasserstein distance $W_2(\mu, \nu)$ measures how different cumulative distribution functions (CDFs) $\mu, \nu$ are from each other by asking how we can transport the mass in $\mu$ to $\nu$ in the most cost-effective manner \citep{panaretos2019statistical}. When distributions are one-dimensional (the focus of our work), the most efficient way of transporting mass between distributions is through their quantiles: $W_2(\mu, \nu) = \left(\int_0^1 \| \mu^{-1}(q) - \nu^{-1}(q) \|^2 dq\right)^{\frac{1}{2}},$ where $\mu^{-1}(q) = \inf\{ x \in \mathbb{R}: \mu(x) \geq q \} \forall q \in [0,1]$ represents the quantile function of $\mu$. The quantile function returns the value $x$ such that the probability of observing a value less than $x$ is at least as much as $q$, the given input probability. Additionally, we can ``average'' distributions using the Wasserstein distance, referred to as barycenters. Specifically, the Wasserstein barycenter of a set of distributions is the distribution that minimizes the average distance between it and all distributions in the set -- the centroid of the set: $\mathbb{B}[F_Y] \in \arg\min_{\gamma}\mathbb{E}[W_2(F_{Y_i}, \gamma)].$
With continuous, one-dimensional distributions, the quantile function of the Wasserstein barycenter also has a closed form solution: $\mathbb{B}[F_{Y}]^{-1}(q) = \mathbb{E}[F_{Y}^{-1}(q)].$ In other words, the quantile function of the average of distributions is the average of quantile functions. We exploit this geometry and represent distributional data via quantiles.

In our setting, we observe $\mathcal{S}_n,$ a collection of $n$ independent and identically distributed observations. Each unit $i$ in $\mathcal{S}_n$ is assigned to a binary treatment $T_i \in \{0,1\}$; for notational convenience, let $\mathcal{S}_n^{(t)}$ represent the set of units whose assigned treatment is $t$. We let $F_{Y_i(1)}$ and $F_{Y_i(0)}$ represent the treated and control potential outcomes, respectively. We make the standard Stable Unit Treatment Value Assumption (SUTVA); specifically, let $F_{Y_i} = F_{Y_i(1)}$ if $t_i = 1$ and $F_{Y_i} = F_{Y_i(0)}$ if $t_i = 0$ \citep{rubin2005causal}. Unlike traditional causal inference that assumes the outcomes exist in some Euclidean space, we consider the setting in which the outcomes are continuous distribution functions in the 2-Wasserstein metric space on the closed interval $\mathcal{I} = [\zeta_{\min}, \zeta_{\max}] \subset \mathbb{R},$ denoted as $\mathcal{W}_2(\mathcal{I}).$ For any cumulative distribution function $F_Y \in \mathcal{W}_2(\mathcal{I}),$ $F_Y(s) = 0$ for all $s \leq \zeta_{\min}$ and $F_Y(s) = 1$ for all $s \geq \zeta_{\max}.$ Additionally, let $F_{\bm{X}_i} = [F_{x_{i,1}}, \ldots, F_{x_{i,d}}]$ represent a vector of $d$ \textit{distributional} covariates for unit $i$ with supports contained within the compact set $\mathcal{J} \subset \mathbb{R}.$ 
\begin{remark}
    Because any scalar can be represented as a degenerate distribution, ADD MALTS can handle distribution-on-scalar, scalar-on-distribution, scalar-on-scalar, \textit{and} distribution-on-distribution regression.
\end{remark}

Similar to \citet{lin2023causal} and \citet{gunsilius2023distributional}, we measure the treatment effect as a contrast between the quantile functions of the potential outcomes. Specifically, we define the individual treatment effect (ITE) as
$\tau_i(q) = F_{Y_i(1)}^{-1}(q) - F_{Y_i(0)}^{-1}(q)$ for all $q \in [0,1].$ We then define the conditional average treatment effect (CATE) and average treatment effect (ATE) by averaging ITEs: respectively, $\tau(q | F_{\bm{x}}) = \mathbb{E}[\tau_i(q) | F_{\bm{X}_i} = F_{\bm{x}}]$ and $ \tau(q) = \mathbb{E}[\tau_i(q)]$ for all $q \in [0,1],$ where the expectations are over the observed population.

\begin{remark}
This estimand is different than the quantile treatment effect studied in scalar causal inference \citep{lin2023causal}: while quantile treatment effects measure the distribution of differences between potential outcomes, our treatment effect measures the difference between distributional potential outcomes.
\end{remark}

We consider the setting where each unit's treatment assignment and the observed potential outcome may depend on common covariates, referred to as confounders. 
Under the following assumptions, we can identify (conditional) average treatment effects. First, we assume conditional ignorability, i.e., that the potential outcomes are independent of the assigned treatment given the confounders: $(F_{Y_i(1)}, F_{Y_i(0)}) \perp \!\!\! \perp T_i \mid F_{\bm{X}_i}.$ Additionally, we assume positivity, i.e., that every unit could be in the treated/control group with some chance: $0 < \mathbb{P}(T_i = 1 \mid F_{\bm{X}_i} = F_{\bm{x}}) < 1.$ Under these assumptions, we can identify CATEs/ATEs (see Proposition \ref{prop:identification} in Section \ref{sec:identificationATE} of the supplement).

\subsection{Related Literature}

\citet{lin2023causal} present three strategies for estimating these treatment effects: outcome regression, propensity score weighting, and a doubly robust approach. One outcome regression scheme is to treat the distributional outcome as functional and use functional data analysis tools \citep{morris2015functional}. Similarly, another approach is to predict each quantile of the outcomes using a separate regression \citep{lin2023causal}. However, neither of these approaches can guarantee that the predicted distributional outcome is actually a \textit{distribution}, i.e., integrates to one, with quantile function monotonically increasing. Without these constraints, the imputed counterfactual may not be a distribution.

Other regression approaches combine traditional statistical ideas and take advantage of the linearity of Wasserstein space for univariate distributions via quantile functions. For example, \citet{petersen2019frechet, ghodrati2022distribution} generalize linear models. \citet{ghosal2023shape, chen2021wasserstein, yang2020random} adapt spline methods. \citet{tang2023wasserstein} introduce an expectation-maximization style algorithm. And \citet{qiu2022random} adapt tree algorithms. However, these outcome regression methods are highly sensitive to model misspecification.

\citet{lin2023causal} propose an augmented inverse propensity weighting style method that requires only one of the propensity score or outcome regression models to be correctly specified. However, these approaches do not allow for any type of meaningful validation of the important causal assumptions. For example, violations of the positivity assumption significantly reduce the precision of our treatment effect estimates. In order to validate the positivity assumption, researchers often prune observations that have extremal estimated propensity scores \citep{stuart2010matching, crump2009dealing}. As we demonstrate in Section \ref{sec:overlapDiscovery}, this strategy is incapable of validating this assumption when the propensity score model is incorrectly specified.

Our approach extends the family of Almost Matching Exactly (AME) methods to the setting of distributional data \citep{diamond2013genetic, dieng2019interpretable, parikh2022malts, lanners2023feature, morucci2023matched}. AME methods learn a distance metric in the covariate space in order to group units that are similar on important covariates; in doing so, we create localized balance, overcoming confounding and enabling us to estimate treatment effects. The conceptual simplicity of these methods makes them easily interpretable and accessible to non-technical audiences. Furthermore, as shown in \citet{parikh2023effects}, AME methods can also easily integrate qualitative analyses, better aiding decision-making. We extend the family of AME techniques to the setting of distributional data. Our method is highly flexible, end-to-end interpretable, accurate at CATE estimation, and useful in answering important questions using wearable devices, sensors, and other distributional data.

\section{\MakeUppercase{Methods}} \label{sec:framework}
\paragraph{Distance Metric with Distributional Covariates}
AME methods learn a distance metric in the covariate space to ensure that matched units are most similar on important features. We first extend the notion of a distance metric to the setting of distributional covariates. Let $d_{\mathcal{M}}$ represent a distance metric parameterized by the $d \times d$ diagonal matrix $\mathcal{M}$. We measure the distance between unit $i, j$'s covariates as
\begin{align*}
    d_\mathcal{M}(F_{\bm{x}_i}, F_{\bm{x}_j}) = \sum_{l = 1}^d \mathcal{M}_{l,l}W_2^2\left(F_{x_{i,l}}, F_{x_{j,l}} \right).
\end{align*}

\begin{remark}
Continuous values can be represented as degenerate distributions; and discrete values can be one-hot-encoded and then represented as degenerate distributions. Because the 2-Wasserstein distance between degenerate distributions is the same as the $\ell_2$ distance between their scalar counterparts, this distance metric can be viewed as a distributional generalization of a weighted Euclidean distance.
\end{remark}

\paragraph{Distance Metric Learning} Traditional nearest neighbor/caliper matching techniques run into the curse of dimensionality where the distance metric can be dominated by less useful covariates when there are many of them \citep{diamond2013genetic}. Instead, we \textit{learn} a distance metric and overcome these issues. We first split our data into a training set $\mathcal{S}_{tr}$ and an estimation set $\mathcal{S}_{est}$; our training and estimation sets are disjoint, so our causal inference remains ``honest'' and helps lower bias \citep{rubin2005causal, athey2016recursive}. On our training set, we evaluate the performance of a proposed distance metric by measuring how well we can predict the observed outcomes; we predict the outcomes for each training unit by averaging the quantile functions of the K-nearest neighbors (KNN) that have the same treatment, where the learned distance metric defines the nearest neighbors. 

Under a distance metric $d_\mathcal{M},$ the KNN of an unit $i$ in the set of observations with treatment $t$, $\mathcal{S}^{(t)},$ is
\begin{align} \label{eqn:knn}
    &KNN_{d_\mathcal{M}}\left(F_{\bm{x}_i}, \mathcal{S}^{(t)} \right) \\
    &=\left\{k:\sum_{j\in \mathcal{S}^{(t)}} \mathbf{1}
                \begin{bmatrix}
                d_{\mathcal{M}}(F_{\bm{x}_i},F_{\bm{x}_j})\;\;\\
                \;\; < d_{\mathcal{M}}(F_{\bm{x}_i},F_{\bm{x}_k})
                \end{bmatrix}
                < K\right\}. \notag
\end{align}
We predict unit $i$'s outcome by computing the quantile function of the barycenter of their KNNs' outcomes:
\begin{align} \label{eqn:fhat}
    \hat{F}_{Y_i}^{-1}(q) = \frac{1}{K}\sum\nolimits_{j \in KNN_{d_\mathcal{M}}(F_{\bm{x}_i}, \mathcal{S}^{(t)})} F_{Y_j}^{-1}(q).
\end{align}

We then find the optimal distance metric parameters that would yield the best predictions of the observed outcomes using the following objective, a distributional generalization of the mean squared error:
\begin{align}
    &\mathcal{M}^*(\mathcal{S}_{tr}) \in \arg\min_{\mathcal{M}}c\|\mathcal{M}\|_{Fr} + \Delta^{(1)}(\mathcal{M}) + \Delta^{(0)}(\mathcal{M}), \notag \\
    &\text{where } \Delta^{(t)} = 1/|\mathcal{S}_{tr}^{(t)}| \sum\nolimits_{i \in \mathcal{S}_{tr}^{(t)}}  W_2^2( \hat{F}_{Y_i}, F_{Y_i} ). \label{eqn:loss}
\end{align}
Our objective function regularizes the parameters using the Frobenius norm and also considers two treatment-specific loss functions. We evaluate how well we can predict the observed outcomes in the training data by calculating the mean squared Wasserstein distance between the predicted and observed values for the treated units and then the control units.

\paragraph{CATE Estimation} On the estimation set, we then estimate treatment effects via matching. Specifically, we estimate the quantile functions of the treated and control conditional barycenter for each treatment $t$ using the set of KNNs: $\hat{\mathbb{B}}[F_Y | F_{\bm{X}} = F_{\bm{x}_i}, T = t]^{-1}(q) = \frac{1}{K}\sum_{j \in KNN_{d_\mathcal{M}}(F_{\bm{x}_i}, \mathcal{S}_{est}^{(t)})} F_{Y_j}^{-1}(q).$ We then estimate the CATE as the difference between the conditional barycenters' quantile functions: 
\begin{align*}
\hat{\tau}(q|F_{\bm{x}_i}) = &\hat{\mathbb{B}}[F_Y | F_{\bm{X}} = F_{\bm{x}_i}, T = 1]^{-1}(q) \\
- &\hat{\mathbb{B}}[F_Y | F_{\bm{X}} = F_{\bm{x}_i}, T = 0]^{-1}(q).
\end{align*}

\subsection{Theoretical Results}
We prove that ADD MALTS consistently estimates conditional barycenters and CATEs by making assumptions analogous to standard ones in the matching literature. 
Assumption \ref{assm:smoothness} is a Lipschitz continuity-style assumption that states that as the units' covariates become more similar, so too do their conditional barycenters. This is an extension of standard assumptions in the matching literature  \citep{dieng2019interpretable, parikh2022malts, lanners2023feature} to the setting of distributional data and is guaranteed to hold with distributional covariates/outcomes with bounded supports. 

\begin{assumption} \label{assm:smoothness}
    Let $F_{\bm{x}_i}, F_{\bm{x}_j} \in \mathcal{W}_2(\mathcal{J})$ and assume $t_i = t_j.$ If $d_\mathcal{M}(F_{\bm{x}_i}, F_{\bm{x}_j}) < \alpha$ for some $\alpha \in \mathbb{R},$ then $W_\infty(\mathbb{B}[F_{Y} | F_{\bm{X}} = F_{\bm{x}_i}, T = t_i], \mathbb{B}[F_{Y} | F_{\bm{X}} = F_{\bm{x}_j}, T = t_j] ) < \delta(\alpha)$ for some monotonically increasing, zero-intercept function $\delta,$ where $W_\infty$ represents the $\infty-$Wasserstein distance.
\end{assumption}

Note that this assumption is under the $\infty$-Wasserstein metric, which  is measured as the largest pointwise distance between quantiles: $W_{\infty}(\mu, \nu) = \sup_{q \in [0,1]}|\mu^{-1}(q) - \nu^{-1}(q)|.$ The $W_\infty$ upper bounds all other $p$-Wasserstein distances (Remark 6.6 in \citet{villani2009optimal}), so this assumption suggests that conditional barycenters become more similar with respect to \textit{any} Wasserstein metric as the covariates become more similar.

The following lemma and theorem rely on this assumption to prove consistency with an intuitive argument. Lemma 1 shows that as we increase the amount of observed data, the radius of each KNN set will decrease. As the radius of each KNN set decreases, the average of the KNN's outcomes will become more similar to that of the query unit's conditional barycenter. This yields the result in Theorem 1: as the estimated conditional barycenters converge to the true conditional barycenters, so too will the estimated CATEs.

\begin{lemma} \label{lemma:barycenter_consistency}
Let Assumption \ref{assm:smoothness} hold. Let $\hat{\mathbb{B}}[F_{Y} | F_{\mathbf{X}} = F_{\mathbf{x}_i}, T = t] \in \underset{\gamma \in \mathcal{W}_2(\mathcal{I})}{\arg\min} \frac{1}{K}\sum_{k = 1}^K W_2^2(F_{Y_k}, \gamma)$ be the barycenter of the KNN's outcomes. Assume the quantile functions of the distributions $F_{Y} \in \mathcal{W}_2(\mathcal{I})$ are Lipschitz continuous in the probability: there exists an $L_F > 0$ such that $|F_Y^{-1}(q) - F_Y^{-1}(q')| \leq L_F|q - q'|$ for $q, q' \in [0,1].$ Let $\varepsilon > \delta(\alpha)$, where $\alpha$ represents the distance from unit $i$ to its $K^{th}$ nearest neighbor according to $d_\mathcal{M}$. Then, there exists a function $c(\varepsilon, \alpha)$ that is exponentially decreasing to 0 in $K$ such that
        \begin{align*}&\mathbb{P}\left(W_1\begin{pmatrix}\hat{\mathbb{B}}[F_{Y} | F_{\mathbf{X}} = F_{\mathbf{x}_i}, T = t],\\ 
                    \mathbb{B}[F_{Y} | F_{\mathbf{X}} = F_{\mathbf{x}_i}, T = t] \end{pmatrix} > \varepsilon \right) \leq 2c(\varepsilon, \alpha),
        \end{align*}
        where $W_1$ is the 1-Wasserstein metric.
\end{lemma}

\begin{theorem} \label{thm:cate_consistency}
    Under the same conditions as Lemma \ref{lemma:barycenter_consistency},
    \begin{align*}
        \mathbb{P}\left(\int_0^1 \left|\tau(q | F_{\mathbf{x}_i}) - \hat{\tau}(q | F_{\mathbf{x}_i}) \right| dq > \varepsilon \right) \leq 4c\left(\frac{\varepsilon}{2},\alpha\right).
    \end{align*}

\end{theorem}

As evident in Theorem \ref{thm:cate_consistency}, for any $\varepsilon > \delta(\alpha)$, the right hand side will decrease to 0 as $K \to \infty.$ Because we work on a bounded covariate space, $\alpha$ is guaranteed to decrease and $\delta(\alpha)$ will go to 0 as the size of the estimation set increases. Therefore, we can consistently estimate conditional barycenters and conditional average treatment effects.

\section{\MakeUppercase{Simulation Experiments}} \label{sec:sims}

Our experiments investigate elements essential for causal inference with distributional data: accuracy and trustworthiness. Section \ref{sec:cateEstimation} shows that ADD MALTS estimates CATEs more accurately than baselines and illustrates that ADD MALTS handles scalar \textit{and} distributional covariates. Section \ref{sec:overlapDiscovery} highlights that ADD MALTS can assess positivity violations. 

\subsection{CATE Estimation} \label{sec:cateEstimation}

\begin{figure*}[t]
    \centering
    \includegraphics[width=0.75\textwidth]{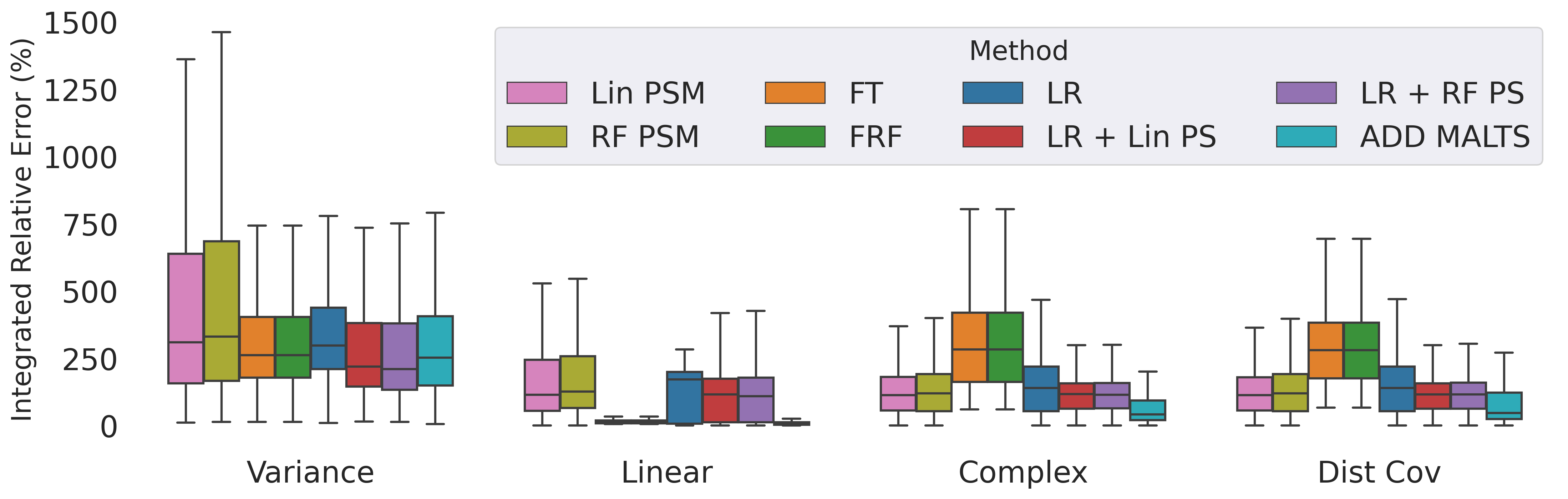}
    \caption{The figure displays the Integrated Relative Error (\%) (y-axis) of the different methods we consider for different simulation setups (x-axis). We consider the following baseline methods: \textbf{Lin PSM} and \textbf{RF PSM} represent propensity score matching fit with linear and random forest models, respectively; \textbf{FT} and \textbf{FRF} represent decision tree and random forest methods for functional outcomes \citep{qiu2022random}; \textbf{LR} represents outcome regression fit at each quantile with a linear regression \citep{lin2023causal}; \textbf{LR + Lin PS} and \textbf{LR + RF PS} represent augmented inverse propensity weighting methods combining the linear outcome regression with linear and random forest propensity score models, respectively.}
    \label{fig:cate_estimation_collated}
\end{figure*}

Our first experiment evaluates how well a variety of baselines and ADD MALTS can estimate CATEs. We consider four data generative processes (DGPs). In each DGP, we generate our distributional outcomes as truncated normal distributions (truncated at $\pm 3$ standard deviations from the mean). In the ``Linear,'' ``Variance,'' and ``Complex'' DGPs, we sample \textit{scalar} covariates while the ``Dist Cov'' DGP has scalar covariates and one \textit{distributional} covariate. A summary of the distributional covariate is used to generate the distributional outcome in ``Dist Cov.'' 

We evaluate each methods ability to estimate the CATE, $\tau(q|F_{\bm{x}_i})$, using the percent Integrated Relative Error: $IRE=100 \times \int_0^1 \left| \frac{\hat{\tau}(q) - \tau(q)}{\tau(q)} \right|dq.$ Section ``CATE Estimation Experimental Details'' of the supplement expands on our experimental setup.

Figure \ref{fig:cate_estimation_collated} displays the results of our simulations. The y-axis represents the integrated relative error so smaller values mean better performance. Across the board, ADD MALTS performs at least as well as the baselines (described in the caption). The ``Complex'' DGP has trigonometric and polynomial relationships between covariates and distributional outcomes. \textbf{In this complex setting, ADD MALTS achieves a median IRE of 41.7\%, approximately one-third of the error of the next best method} (LR + RF PS with median IRE of $\approx 115\%$).

In the ``Dist Cov'' DGP, the specific function that summarizes the distributional covariate is the integral over the covariate's quantile function (see the table in ``CATE Estiamtion Experimental Details'' of the supplement). However, in practice, the correct summary function is unknown (e.g., using the mean, median, area under the quantile function). Here, we advantage the baseline methods by letting them use the correct summary value of the distribution (i.e., the integral over its quantile function). On the other hand, ADD MALTS only has access to the raw data drawn from the distribution. \textbf{Even in this scenario, ADD MALTS outperforms the baselines}.
ADD MALTS again reduces the median IRE from the next best method by one-third, demonstrating that \textbf{ADD MALTS can effectively handle both scalar and distributional covariates}. ADD MALTS does not require us to preprocess or summarize distributional covariates, proving that ADD MALTS handles complex data without sacrificing performance.

\subsection{Positivity Violations} \label{sec:overlapDiscovery}

\begin{figure}[t]
    \centering
    \includegraphics[width = .42\textwidth]{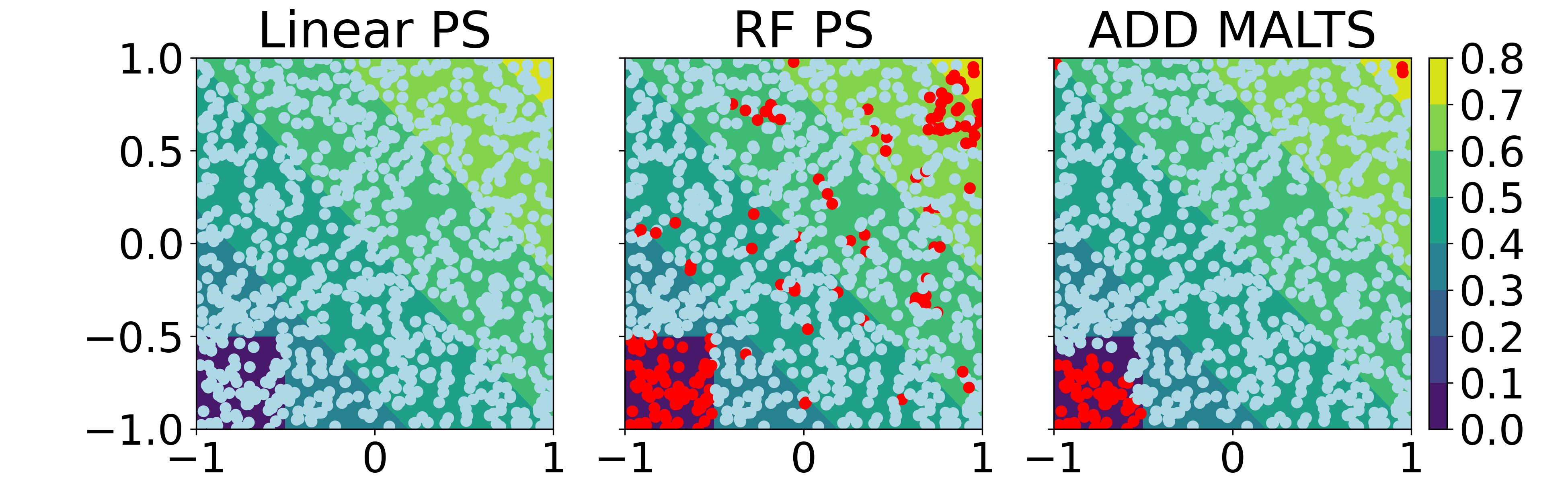}
    \caption{The plot displays which units should be pruned in red according to each method: (from left to right) propensity score estimated with logistic regression using L1 regularization, propensity score estimated with a random forest, and the diameter of matched groups estimated with ADD MALTS. The background displays the true propensity score; the bottom, left corner marks the region of the covariate space with no overlap.}
    \label{fig:overlap}
\end{figure}

ADD MALTS can also precisely assess positivity violations. While there are several methods for characterizing regions with violations of positivity with scalar data \citep[e.g.,][]{oberst2020characterization, crump2009dealing, hill2020bayesian}, to the best of our knowledge, there are no techniques for distributional data. To benchmark ADD MALTS, we extend methods using estimated propensity scores to our setting. Traditionally, researchers would exclude any units for which the estimated propensity score is not within certain thresholds (e.g., between 0.1 and 0.9) \citep{stuart2010matching, crump2009dealing, li2019addressing}. However, these approaches are highly sensitive to model misspecification. We demonstrate that ADD MALTS assesses positivity violations more accurately than these propensity score baselines.

\paragraph{Using ADD MALTS to Assess Overlap.} First, we match each treated and control unit to their K nearest neighbors of the opposite treatment status. We calculate each unit's nearest-neighbor diameter $D_i$, the average distance to its nearest neighbors: $D_i = \frac{1}{K}\sum_{j \in KNN_{d_{\mathcal{M}}}}d_{\mathcal{M}}(F_{\bm{x}_i}, \mathcal{S}^{1-t_i}).$ Units with high diameters are far away from units of the opposite treatment and are therefore more likely to be in regions of the covariate space with limited overlap. We flag any units whose diameter is greater than $D_{\text{upper}} = Q_{D_i}(0.75) + 1.5 \cdot (Q_{D_i}(0.75) - Q_{D_i}(0.25)),$ where $Q_{D_i}(s)$ is the $s^{th}$ quantile of the diameters (this is a common measure of outliers \citep{suri2019outlier}).

\paragraph{Simulation Setup} We simulate data using the following DGP. We have two covariates $x_{i,0}, x_{i,1} \sim \text{Unif}[-1,1]$ and the following, piece-wise propensity score model: $T_i = 0$ if $x_{i,0} \leq -0.5 \land x_{i,1} \leq -0.5,$ else $T_i \sim \text{Bern}( \text{expit}(-0.5x_{i,0} -0.5x_{i,1})).$ In this simulation setup, the true propensity score is linear except for the bottom left corner of the covariate space, where there is no overlap. We train a (parametric) $\ell_1$-regularized linear propensity score model (Linear PS) and a (non-parametric) random forest propensity score model (RF PS); any units with propensity scores outside $[0.1, 0.9]$ were labeled as suffering from positivity violations. 

Figure \ref{fig:overlap} displays the results for one of the iterations of this simulation. The linear propensity score (Linear PS) fails to flag positivity violations. The random forest propensity score (RF PS) correctly characterizes the region of space with positivity violations but at the cost of mischaracterizing regions of overlap as having positivity violations. Dropping observations in regions of the space without positivity violations could adversely affect the precision of our treatment effect estimates. In contrast, ADD MALTS characterizes almost all of the regions of the covariate space correctly. We repeat this experiment 100 times and find that -- overall -- ADD MALTS accurately classifies 97.6\% of units. In comparison, Linear PS and RF PS only classify 93\% of units properly. ADD MALTS also enables us to inspect nearest neighbor sets and qualitatively assess whether flagged units suffer from a positivity violation, unlike the propensity score methods. \textbf{ADD MALTS precisely flags overlap violations while also being end-to-end interpretable.}

\section{\MakeUppercase{Real Data Analysis}} \label{sec:cgm}
We use ADD MALTS to reanalyze a clinical trial \citep{juvenile2008continuous} focused on assessing the effectiveness of continuous glucose monitors (CGMs) in mitigating the risk of hyperglycemia (resulting from high glucose levels) or hypoglycemia (linked to low glucose levels) in type 1 diabetes patients. 
To ensure data validity, we begin by investigating potential violations of the positivity assumption in the CGM trial data. ADD MALTS detects a lack of positivity for patients who experienced severe hypoglycemia prior to treatment, raising concerns about the generalizability of the trial results to this subgroup. We then assess the heterogeneity of the treatment effects using ADD MALTS. Our findings suggest that, on average, the use of CGMs provides only a marginal benefit in reducing the risk of hyperglycemia or hypoglycemia. However, our subgroup analysis reveals that \textbf{CGMs can be beneficial in reducing the risk of extremely high glucose levels for patients aged 55 or older who are effectively managing their diabetes.}

\paragraph{Data Description} CGMs are wearable devices that monitor blood glucose levels. The Juvenile Diabetes Research Foundation (JDRF) conducted a randomized control trial across 10 clinics and a cohort of 450 patients with type 1 diabetes to assess how helpful CGMs can be in mitigating the risk of extremal glucose concentrations \citep{juvenile2008continuous, juvenile2010effectiveness}. One week prior to randomization, all patients wore modified CGMs where the readings were recorded but not visible to diabetes patients. Patients were then randomly assigned to monitor blood glucose concentrations using CGMs (treatment) or a standard, blood glucose meter (control). The researchers used a stratified randomization scheme to maintain balance based on the clinical center, age group (8 to 14 years, 15 to 24 years, and $\geq 25$ years), and baseline blood glycated hemoglobin levels ($HbA1c \leq$ or $> 8\%$). Patients monitored their blood glucose levels using their assigned strategy for 26 weeks; after 26 weeks, all patients wore CGMs with the readings blinded to the control group and visible to the treated group. 

\begin{figure}
    \centering
    \includegraphics[width=0.42\textwidth]{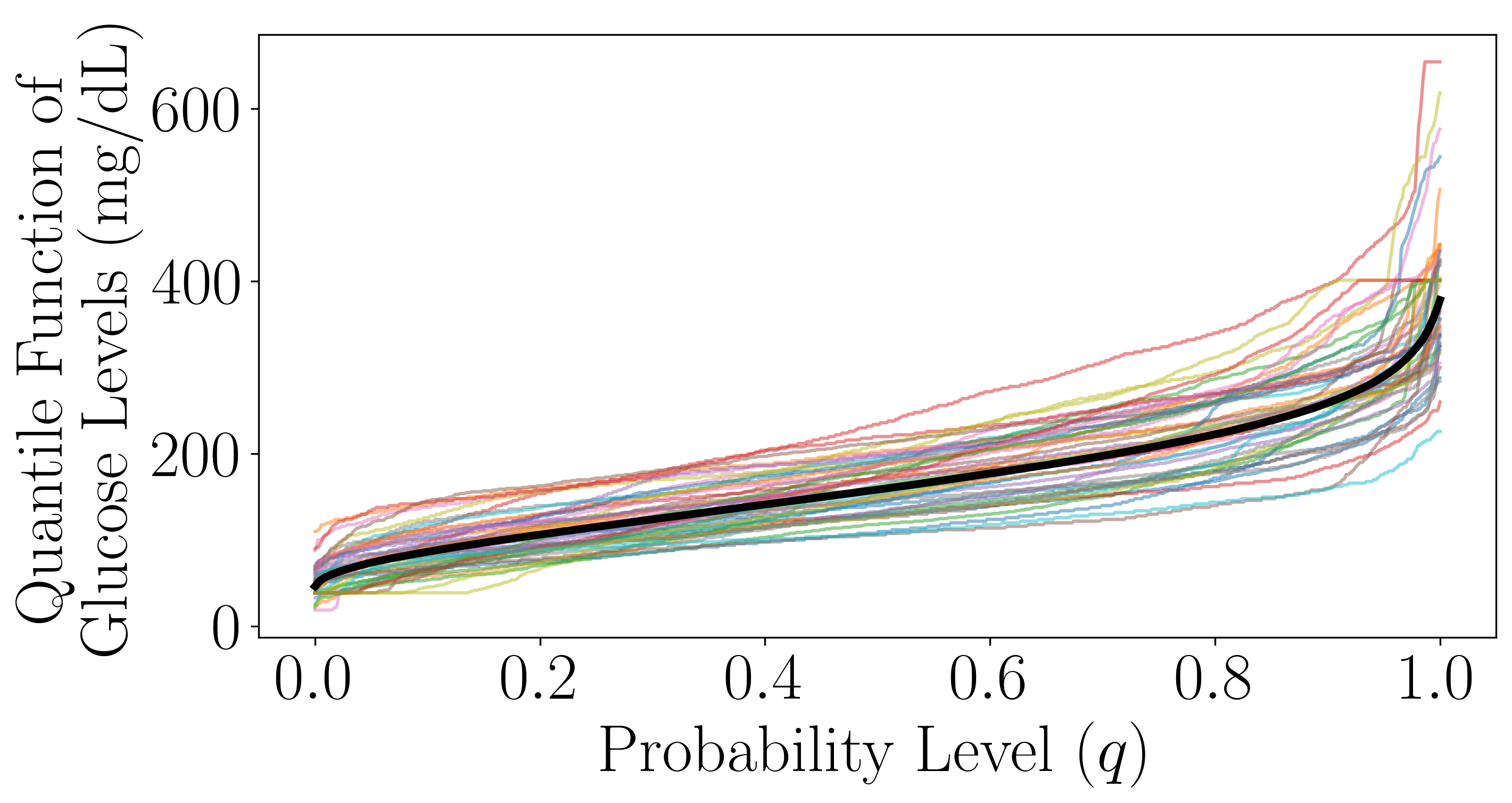}
    \caption{Quantile functions of glucose levels measured at baseline. The thick black line represents the \textit{average} quantile function while the other colors represent the quantile functions for 50 patients.}
    \label{fig:baseline_quantile_fns}
\end{figure}

\paragraph{Methods} Previous analyses of these data use the time in range metric to summarize CGM readings \citep{juvenile2008continuous, juvenile2010effectiveness}. As we demonstrate in Figure \ref{fig:bounds_effect_lower70}, the time in range metric 
is highly sensitive to the actual choice of healthy range. To overcome these issues, we re-analyze this data by representing each patient's CGM data as distributions of glucose concentrations over time (Figure \ref{fig:baseline_quantile_fns} shows the quantile functions of the baseline glucose distributions for 50 patients). We assess overlap and estimate the ATE and CATEs using ADD MALTS.

In our analysis, we include the following control variables. Age describes the unit's age at randomization (in years). HbA1c describes whether the unit's glycated blood hemoglobin at baseline was $\leq8\%$ (Low) or not (High). ``Dur'' describes the number of years the patient had diabetes. ``Col'' denotes whether the patient (or their guardian) graduated from college. ``NHW?'' is a boolean that is true if the patient is Non-Hispanic White. ``Hypo?'' denotes whether the patient suffered from an episode of severe hypoglycemia prior to treatment. ``Male'' denotes whether the patient is male. And Treatment denotes the patient's treatment assignment. We also control for the distribution of pre-treatment glucose concentrations, as measured by blinded CGMs. We use 40\% of the patients to learn ADD MALTS' distance metric and 60\% of the patients to estimate treatment effects.

\paragraph{Assessing Positivity Violations} We first check for positivity violations in our estimation set. We flagged units in no-overlap regions using the same procedure as in Section \ref{sec:overlapDiscovery}. Table \ref{tab:hypoPatient} displays the nearest neighbor set for a 43 year old male in the control group who suffered from severe hypoglycemia (i.e., an adverse health outcome due to glucose levels being too low) who was flagged. When we inspect his nearest neighbor set, we see that \textbf{there is no similar treated patient who also suffered an episode of severe pre-treatment hypoglycemia}. This unit's CATE is not very trustworthy, and we would need more data to make such granular insights. 

\begin{table}[t]
    \centering
    \resizebox{\linewidth}{!}{
    \begin{tabular}
    {|c|c|c|c|c|c|c|c|c|}
\hline 
\textbf{Age} &  \textbf{HbA1c} &  \textbf{Dur.} & \textbf{Col} &  \textbf{NHW?} &  \textbf{Hypo?} &  \textbf{Male} &  \textbf{T} \\
\hline
             43 &                Low &         24.3 &            True &              True &     \textbf{True} &  True &          0 \\\hline
             42 &               High &         20.3 &            True &              True &    \textbf{False} &  True &          1 \\
             40 &               Low &         24.2 &            True &              True &    \textbf{False} & False &          1 \\
              43 &                High &         20.8 &            True &              True &    \textbf{False} & False &          1 \\
             40 &               Low &         33.1 &            True &              True &    \textbf{False} & False &          1 \\
              41 &               Low &         16.0 &            True &              True &    \textbf{False} &  True &          1 \\
\hline
\end{tabular}}
    \caption{The table displays the treated nearest neighbors (bottom five rows) for the query unit (top row).}
    \label{tab:hypoPatient}
\end{table}

\paragraph{Estimated Treatment Effects} 

\begin{figure}[t]
    \centering
    \includegraphics[width=0.4\textwidth]{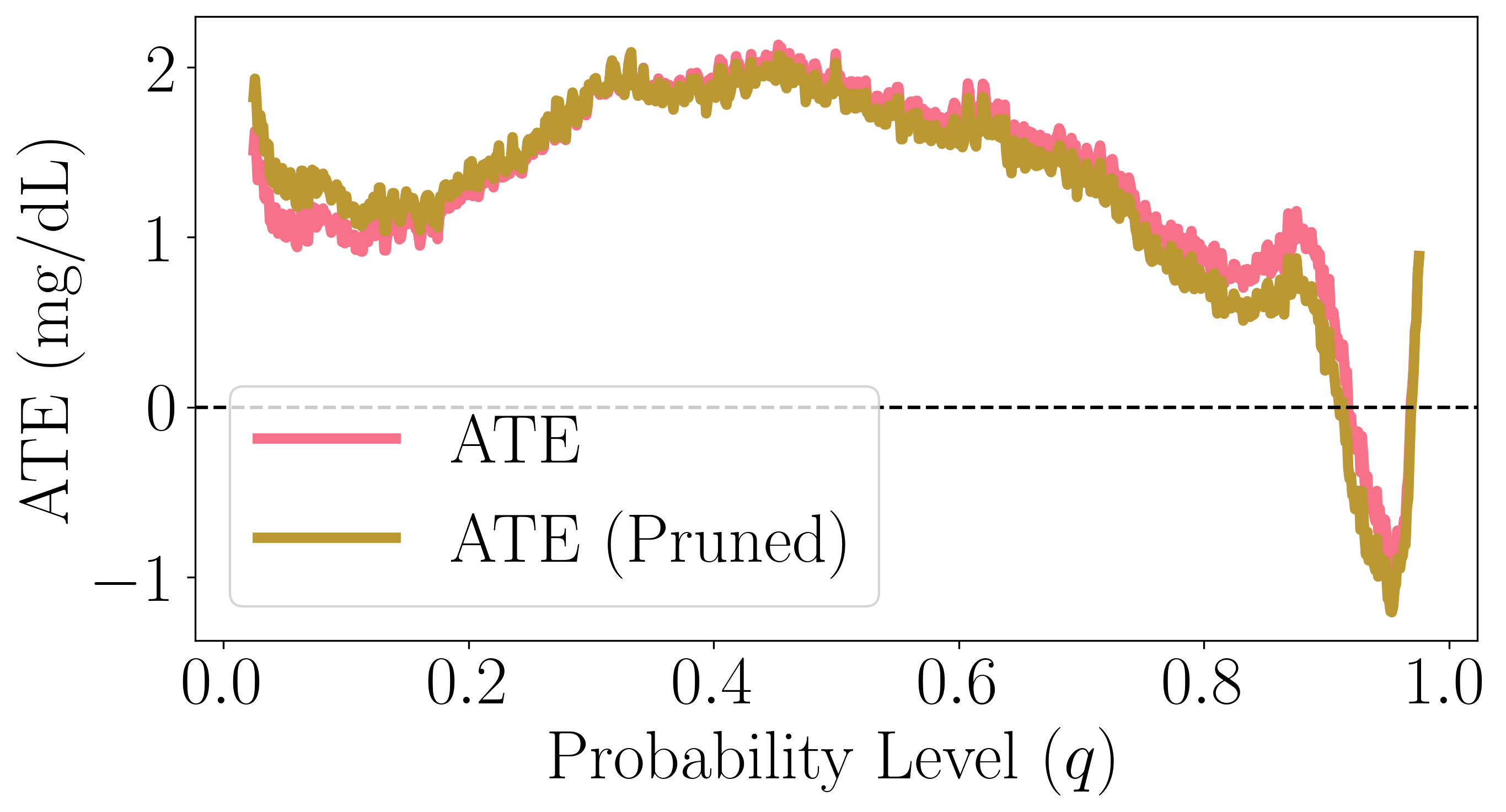} \\
    (a)\\
    \includegraphics[width=0.4\textwidth]{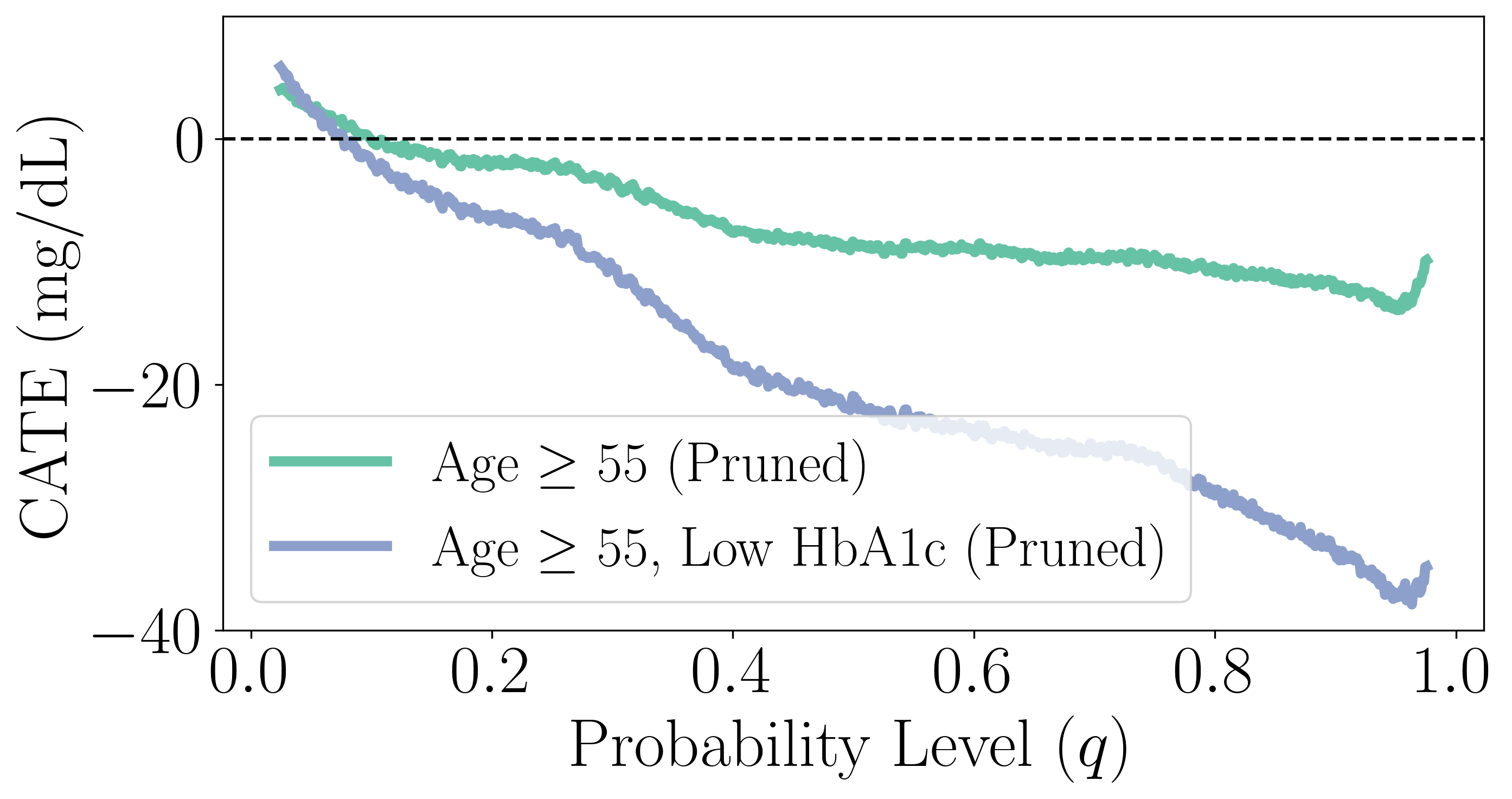}\\
    (b)
    \caption{(a) Average treatment effect of glucose-monitoring with CGM on the distribution of glucose concentrations before pruning positivity violations (pink) and after (gold). (b) Conditional average treatment effect of glucose monitoring with CGM on the distribution of glucose concentrations for patients older than 55 years of age (green) and for those with low HbA1c levels at baseline (blue). The x-axes display the probability level and the y-axes display the difference in the related quantiles of the outcomes.}
    \label{fig:ate}
\end{figure}

We first estimate the average treatment effect. As shown in Figure \ref{fig:ate}, there is little difference between the overall ATE (pink) and the ATE after pruning units suffering from positivity violations (gold). There is a very small, marginal change in glucose concentrations: at each quantile, CGMs only affect glucose levels by between -2 and 1.5 mg/dL, which is miniscule when considering the average person's glucose readings range between 50 and 350 mg/dL (see Figure \ref{fig:baseline_quantile_fns}). \textbf{We find that, on average, CGMs do not affect glucose levels}.

ADD MALTS enables us to go beyond ATEs and accurately investigate effects in subpopulations. Using ADD MALTS, we revisit the subpopulation in Figure \ref{fig:bounds_effect_lower70}: patients older than 55 years of age. As shown in Figure \ref{fig:ate}(b), CGMs have marginal effects on glucose concentrations for patients older than 55 years of age (green line); extremal glucose levels only change by up to 15 mg/dL. As we see in Figure \ref{fig:baseline_quantile_fns}, the average patient's glucose concentrations range between 50 and 350 mg/dL. A decrease in upper-extremal glucose levels by 15 mg/dL suggests CGMs may be beneficial but not transformative for these older adults' hyperglycemic risks.

However, CGMs may be very beneficial for patients older than 55 years old who also have low HbA1c levels at baseline; HbA1c is a biomarker that measures the concentration of long-term sugars in the bloodstream. People with lower HbA1c levels tend to have better control of their diabetes needs and lifestyle (i.e., managing diet and exercise). For these patients, upper extremal glucose levels decrease by up to 40 mg/dL, over 10\% of the maximum glucose concentration of the average patient (see black line in Figure \ref{fig:baseline_quantile_fns}). Because the effect has intensified for patients with low HbA1c levels, this treatment effect estimate suggests that CGMs are most beneficial for patients who actively manage their diabetes needs. On their own, CGMs are not a panacea for diabetes care; however, when coupled with active engagement and self-care, CGMs can help reduce patients' risk of hyperglycemia.

CGMs tend not to increase \textit{lower} extremal glucose levels, suggesting that CGMs may not be effective in mitigating the risk of hypoglycemia. As suggested by \cite{wolpert2007use}, CGMs may cause patients to ``over bolus'' or overcompensate for rising glucose levels if they observe glucose readings too often. Patients overly concerned about high glucose levels may not manage low glucose concentrations. Understanding that CGMs do not increase low glucose levels suggests 
CGMs tend not to increase \textit{lower} extremal glucose levels, suggesting that CGMs may not be effective in mitigating the risk of hypoglycemia. As suggested by \cite{wolpert2007use}, CGMs may cause patients to ``over bolus'' or overcompensate for rising glucose levels if they observe glucose readings too often. Patients overly concerned about high glucose levels may not manage low glucose concentrations. Understanding that CGMs may not change low glucose levels suggests that we should explore other treatments to help diabetic patients better manage hypoglycemic risks.

\section{\MakeUppercase{Conclusion}} \label{sec:conclusion}
To enable high-quality and trustworthy causal inference with distributional data, we introduce ADD MALTS. We prove that ADD MALTS can consistently estimate CATEs and validate its performance via simulation. We also show that ADD MALTS effectively handles distributional \textit{and} scalar covariates. ADD MALTS can also precisely flag overlap violations. We use ADD MALTS to study CGMs' role in glucose level management in type 1 diabetic patients.

\paragraph{Limitations and Future Directions} While we discuss ADD MALTS' utility in the context of continuous glucose monitoring data, it can be highly useful for analyzing a variety of other data. For example, distributional representations have been helpful for summarizing images \citep{oliva2013distribution, yang2020quantile, zhang2022high} and for summarizing survey data from across geographies \citep{gunsilius2023distributional}. However, these complex data may benefit from being represented as \textit{multidimensional} distributions. Future directions should consider extending ADD MALTS to the setting of multidimensional distributional data. 
Additionally, uncertainty quantification and variance estimation using distributional outcomes is a difficult and challenging issue that future research should explore. We propose preliminary insights in the supplementary material. Finally, wearable devices are becoming increasingly popular, leading to large datasets with hundreds of thousands of patients \citep{nazaret2022modeling}. To accommodate such data, future research could benefit from extensions of ADD MALTS that scale to larger datasets without sacrificing accuracy or interpretability.

\section*{\MakeUppercase{Acknowledgements}}
We gratefully acknowledge support from NIH/NIDA R01DA054994, NIH/NIAID R01AI143381,
DOE DE-SC0023194, NSF IIS-2147061, and NSF IIS-2130250. Alexander Volfovsky is also supported by a National Science Foundation Faculty Early Career Development Award (CAREER: Design and analysis of experiments for complex
social processes). National Institute on Drug Abuse (NIDA) R01DA056407 supports Harsh Parikh. The authors want to thank Quinn Lanners for his insightful and constructive comments.

\bibliographystyle{apalike}
\bibliography{references}

\begin{appendices}
    
\onecolumn

\aistatstitle{Interpretable Causal Inference for Analyzing Wearable, Sensor, and Distributional Data: Supplementary Materials}

\section{Identification of the ATE} \label{sec:identificationATE}
We first prove that we can identify average treatment effects. This is the same theorem statement as in \citet{lin2023causal} and the proof follows the same arguments. This is here for completeness.

\begin{proposition}[\citet{lin2023causal}] \label{prop:identification}
Under SUTVA, conditional ignorability, and positivity, we can identify average treatment effects.
\end{proposition}

\begin{proof}
    We aim to identify the average treatment effect,
    \begin{align*}
        \tau(q) = \mathbb{E}\left[F_{Y_i(1)}^{-1}(q) - F_{Y_i(0)}^{-1}(q) \right],
    \end{align*}
    where the expectation is over the population we sample data from.

    Let $q \in [0,1].$ By the law of iterated expectations,
    \begin{align*}
        \mathbb{E}_i[F_{Y_i(1)}^{-1}(q) - F_{Y_i(0)}^{-1}(q)] 
        &= \mathbb{E}_{F_{\bm{X}}}\left\{ \mathbb{E}_i[F_{Y_i(1)}^{-1}(q) - F_{Y_i(0)}^{-1}(q) |  F_{\bm{X}_i} = F_{\bm{x}}] \right\} \\
        &= \mathbb{E}_{F_{\bm{X}}}\left\{ \mathbb{E}_i[F_{Y_i(1)}^{-1}(q) |  F_{\bm{X}} = F_{\bm{x}}] - \mathbb{E}_i[F_{Y_i(0)}^{-1}(q) | F_{\bm{X}_i} = F_{\bm{x}} ]\right\},
    \end{align*}
    by the linearity of expectations. Recall that by conditional ignorability, we know that for all $q \in [0,1], \mathbb{E}_i[F_{Y_i(T_i)}^{-1}(q) | F_{\bm{x}_i}] = \mathbb{E}_i[F_{Y_i(T_i)}^{-1}(q) | T_i,  F_{\bm{x}_i}].$ Substituting this equality into our equation, we know that
    \begin{align*}
        &\mathbb{E}_i[F_{Y_i(1)}^{-1}(q) - F_{Y_i(0)}^{-1}(q)] \\
        = &\mathbb{E}_{ F_{\bm{X}}}\left\{ \mathbb{E}_i[F_{Y_i(1)}^{-1}(q) |  F_{\bm{X}_i} = F_{\bm{X}}] - \mathbb{E}_i[F_{Y_i(0)}^{-1}(q) | F_{\bm{X}_i} = F_{\bm{x}}] \right\} \\
        = &\mathbb{E}_{F_{\bm{X}}}\left\{ \mathbb{E}_i[F_{Y_i(1)}^{-1}(q) | F_{\bm{X}_i} = F_{\bm{x}}, T_i = 1] - \mathbb{E}_i[F_{Y_i(0)}^{-1}(q) |  F_{\bm{X}_i} = F_{\bm{x}}, T_i = 0] \right\}.
    \end{align*}
    Because we have now conditioned on observing a specific treatment for our units, we know by SUTVA that $F_{Y_i} = F_{Y_i(T_i)}.$ So,
    \begin{align} \label{eqn:identification}
        &\mathbb{E}_i[F_{Y_i(1)}^{-1}(q) - F_{Y_i(0)}^{-1}(q)] \nonumber \\
        = &\mathbb{E}_{F_{\bm{X}}}\left\{ \mathbb{E}_i[F_{Y_i}^{-1}(q) |  F_{\bm{X}} = F_{\bm{x}}, T_i = 1] - \mathbb{E}_i[F_{Y_i}^{-1}(q) |  F_{\bm{X}} = F_{\bm{x}}, T_i = 0] \right\} \nonumber \\
        = &\mathbb{E}_{ F_{\bm{X}}}\left\{ \mathbb{E}_i[F_{Y_i}^{-1}(q) | F_{\bm{X}} = F_{\bm{x}}, T_i = 1] \right\} - \mathbb{E}_{ F_{\bm{X}}}\left\{\mathbb{E}_i[F_{Y_i}^{-1}(q) | F_{\bm{X}} = F_{\bm{x}}, T_i = 0] \right\} \\
        = &\mathbb{E}_i[F_{Y_i}^{-1}(q) | T_i = 1] - \mathbb{E}_i[F_{Y_i}^{-1}(q) | T_i = 0], \nonumber
    \end{align}
    where the last two steps follow by using the linearity of expectations and reversing the iterated expectation.
\end{proof}

\section{CATE Estimation Consistency Proofs} \label{sec:CATEconsistency}
This section establishes the consistency of ADD MALTS' CATE estimation strategy. We first re-introduce the smoothness assumption used in these proofs:
\begin{assumption*}[\ref{assm:smoothness}] 
    Let $F_{\bm{x}_i}, F_{\bm{x}_j} \in \mathcal{W}_2(\mathcal{J})$ and assume $t_i = t_j.$ If $d_\mathcal{M}(F_{\bm{x}_i}, F_{\bm{x}_j}) < \alpha$ for some $\alpha \in \mathbb{R},$ then $W_\infty(\mathbb{B}[F_{Y} | F_{\bm{X}} = F_{\bm{x}_i}, T = t_i], \mathbb{B}[F_{Y} | F_{\bm{X}} = F_{\bm{x}_j}, T = t_j] ) < \delta(\alpha)$ for some monotonically increasing, zero-intercept function $\delta,$ where $W_\infty$ represents the $\infty-$Wasserstein distance.
\end{assumption*}

\subsection{Proof of Barycenter Consistency (Lemma 1)}
\begin{lemma*}[\ref{lemma:barycenter_consistency}]
Let Assumption \ref{assm:smoothness} hold. Let $\hat{\mathbb{B}}[F_{Y} | F_{\mathbf{X}} = F_{\mathbf{x}_i}, T = t] \in \underset{\gamma \in \mathcal{W}_2(\mathcal{I})}{\arg\min} \frac{1}{K}\sum_{k = 1}^K W_2^2(F_{Y_k}, \gamma)$ be the barycenter of the KNN's outcomes. Assume the quantile functions of the distributions $F_{Y} \in \mathcal{W}_2(\mathcal{I})$ are Lipschitz continuous in the probability: there exists an $L_F > 0$ such that $|F_Y^{-1}(q) - F_Y^{-1}(q')| \leq L_F|q - q'|$ for $q, q' \in [0,1].$ Let $\varepsilon > \delta(\alpha)$, where $\alpha$ represents the distance from unit $i$ to its $K^{th}$ nearest neighbor according to $d_\mathcal{M}$. Then, there exists a function $c(\varepsilon, \alpha)$ that is exponentially decreasing to 0 in $K$ such that
        \begin{align*}&\mathbb{P}\left(W_1\begin{pmatrix}\hat{\mathbb{B}}[F_{Y} | F_{\mathbf{X}} = F_{\mathbf{x}_i}, T = t],\\ 
                    \mathbb{B}[F_{Y} | F_{\mathbf{X}} = F_{\mathbf{x}_i}, T = t] \end{pmatrix} > \varepsilon \right) \leq 2c(\varepsilon, \alpha),
        \end{align*}
        where $W_1$ is the 1-Wasserstein metric.
\end{lemma*}

\begin{proof}

Let $\mathbb{B}_i = \mathbb{B}[F_{Y} | F_{\bm{X}} = F_{\bm{x}_i}, T = t]$ represent the conditional barycenter of the outcome at $F_{\bm{x}_i}$ and treatment $t$. Let $KNN_i = KNN_{d_\mathcal{M}}(F_{\bm{x}_i}, \mathcal{S}_{est}^{(t)})$ represent the K nearest neighbors to unit $i$ in the estimation set with assigned treatment $t.$ We demonstrate that ADD MALTS' estimate -- $\hat{\mathbb{B}}_i = \arg\min_{\gamma \in \mathcal{W}_2(\mathcal{I})} \frac{1}{K}\sum_{k = 1}^K W_2^2(F_{Y_k}, \gamma)$ -- consistently estimates the true conditional barycenter's quantile function. Let $\varepsilon > \delta(\alpha).$

We begin by showing the following: for a given $q \in [0,1],$
\begin{align*}
    \mathbb{P}\left( \left| \mathbb{B}^{-1}_i(q) - \hat{\mathbb{B}}^{-1}_i(q) \right| > \varepsilon \right) \leq 2\exp\left( \frac{-K\left(\varepsilon - \delta(\alpha)\right)}{2(\zeta_{\max} - \zeta_{\min})^2} \right).
\end{align*}

Recall that the quantile function of the barycenter of one-dimensional, continuous distributions is the average of quantile functions, so
\begin{equation*}
    \hat{\mathbb{B}}_i^{-1}(q) = \frac{1}{K}\sum_{k \in KNN_i} F_{Y_k}^{-1}(q).
\end{equation*}

We manipulate the probability statement to bring it into the following form:
\begin{align*}
    &\mathbb{P}\left( \left| \mathbb{B}^{-1}_i(q) - \hat{\mathbb{B}}^{-1}_i(q) \right| > \varepsilon \right) \\
    &= \mathbb{P}\left( \left| \mathbb{B}^{-1}_i(q) - \frac{1}{K}\sum_{k \in KNN_i} F_{Y_k}^{-1}(q) \right| > \varepsilon \right) \text{ by def'n on empirical barycenter } \\
    &= \mathbb{P}\left( \left| \frac{1}{K}\sum_{k \in KNN_i} \left[ \mathbb{B}^{-1}_i(q) - F_{Y_k}^{-1}(q) \right] \right| > \varepsilon \right) \text{ by rearranging the summation} \\
    &= \mathbb{P}\left( \left| \frac{1}{K}\sum_{k \in KNN_i} \left[ \mathbb{B}^{-1}_i(q) - \underbrace{\mathbb{B}_k^{-1}(q) + \mathbb{B}_k^{-1}(q)}_{\text{ (add 0) }}- F_{Y_k}^{-1}(q) \right] \right| > \varepsilon \right) \\
    &= \mathbb{P}\left( \left| \frac{1}{K}\sum_{k \in KNN_i} \left[ \mathbb{B}^{-1}_i(q) - \mathbb{B}_k^{-1}(q) \right] +  \frac{1}{K}\sum_{k \in KNN_i}\left[\mathbb{B}_k^{-1}(q)- F_{Y_k}^{-1}(q) \right] \right| > \varepsilon \right) \text{ by distributing the summation} \\
    &\leq \mathbb{P}\left( \left| \frac{1}{K}\sum_{k \in KNN_i} \mathbb{B}^{-1}_i(q) - \mathbb{B}_k^{-1}(q) \right| +  \left| \frac{1}{K}\sum_{k \in KNN_i} \mathbb{B}_k^{-1}(q)- F_{Y_k}^{-1}(q)  \right| > \varepsilon \right) \text{by triangle inequality} \\
    &\leq \mathbb{P}\left( \frac{1}{K}\sum_{k \in KNN_i}  \left| \mathbb{B}^{-1}_i(q) - \mathbb{B}_k^{-1}(q) \right| +  \left| \frac{1}{K}\sum_{k \in KNN_i} \mathbb{B}_k^{-1}(q)- F_{Y_k}^{-1}(q)  \right| > \varepsilon \right) \text{by triangle inequality}.
\end{align*}

Recall that the $\infty$-Wasserstein distance in one dimension is simply a contrast between quantile functions: $W_\infty(F_A, F_B) = \sup_{q \in [0,1]} |F_A^{-1}(q) - F_B^{-1}(q)|.$ Thus,
\begin{align*}
    \frac{1}{K}\sum_{k \in KNN_i}  \left| \mathbb{B}^{-1}_i(q) - \mathbb{B}_k^{-1}(q) \right| 
    \leq \frac{1}{K}\sum_{k \in KNN_i} \sup_{q \in [0,1]} \left| \mathbb{B}^{-1}_i(q) - \mathbb{B}_k^{-1}(q) \right| 
    = \frac{1}{K}\sum_{k \in KNN_i} W_\infty \left(\mathbb{B}_i, \mathbb{B}_k \right).
\end{align*}

Recall that unit $i$ and unit $k$ are within distance $\alpha$ of each other.  By Assumption~\ref{assm:smoothness}, we then know that $W_\infty\left(\mathbb{B}_i, \mathbb{B}_k \right) \leq \delta(\alpha).$ So,
\begin{align*}
    \frac{1}{K}\sum_{k \in KNN_i} W_2\left(\mathbb{B}_i, \mathbb{B}_k \right) 
    \leq \frac{1}{K}\sum_{k \in KNN_i} \delta(\alpha) 
    = \delta(\alpha).
\end{align*}
Then,
\begin{align*}
    &\mathbb{P}\left( \frac{1}{K}\sum_{k \in KNN_i}  \left| \mathbb{B}^{-1}_i(q) - \mathbb{B}_k^{-1}(q) \right| +  \left| \frac{1}{K}\sum_{k \in KNN_i} \mathbb{B}_k^{-1}(q)- F_{Y_k}^{-1}(q)  \right| > \varepsilon \right) \\
    &\leq \mathbb{P}\left( \delta(\alpha) +  \left| \frac{1}{K}\sum_{k \in KNN_i} \mathbb{B}_k^{-1}(q)- F_{Y_k}^{-1}(q)  \right| > \varepsilon \right) \\
    &= \mathbb{P}\left(\left| \frac{1}{K}\sum_{k \in KNN_i} \mathbb{B}_k^{-1}(q)- F_{Y_k}^{-1}(q)  \right| > \varepsilon - \delta(\alpha) \right).
\end{align*}

By Hoeffding's inequality, we then know that
\begin{align}
    &\mathbb{P}\left( \left| \mathbb{B}^{-1}_i(q) - \hat{\mathbb{B}}^{-1}_i(q) \right| > \varepsilon \right) \notag \\
    &\leq \mathbb{P}\left(\left| \frac{1}{K}\sum_{k \in KNN_i} \mathbb{B}_k^{-1}(q)- F_{Y_k}^{-1}(q)  \right| > \varepsilon - \delta(\alpha) \right) \leq 2\exp\left( \frac{-K\left(\varepsilon - \delta(\alpha) \right)^2}{2(\zeta_{\max} - \zeta_{\min})^2} \right). \tag{Pointwise Bound} \label{eqn:pointwise_bound}
\end{align}

This result demonstrates point-wise consistency at each probability level. Now, we will show consistency of estimating the conditional barycenter simultaneously. 

Assume that all quantile functions, $F_{Y_k} \in \mathcal{W}_2(\mathcal{I})$ are Lipschitz continuous. Define the function
\begin{align*}
    g(q) = \left|\frac{1}{K}\sum_{k \in KNN_i} F_{Y_k}^{-1}(q) - \mathbb{B}_i^{-1}(q) \right|.
\end{align*}
Because the average and difference of Lipschitz functions are Lipschitz, we then know that $g(q)$ is also Lipschitz continuous. Let $L$ be the Lipschitz constant of $g.$ 

Let $B_r(x)$ represent a ball of radius $r \in (0,0.5]$ centered at $x \in [0,1]$. Choose an $r$ and a corresponding grid of points $\Vec{Q} = [q_1, \ldots, q_N] \subset [0,1]$ such that the following are satisfied: (1) $r < \frac{\varepsilon - \delta(a)}{L}$ and (2) $\cup_{b = 1}^N B_r(q_b) = [0,1]$ and $B_r(q_a) \cap B_r(q_b) = \emptyset$ for any $q_a \neq q_b$ in $\Vec{Q}.$ Choose any $q \in \Vec{Q}.$ Because all $q' \in B_r(q)$ are within distance $r$ of $q$, we know that $\sup_{q' \in B_r(q)}|g(q) - g(q')| \leq Lr.$ Then,
\begin{align*}
    \int_{B_r(q)} g(t) dt &= \int_{B_r(q)} \left[g(t) - g(q) + g(q) \right] dt \text{ by adding 0 }\\
&\leq \int_{B_r(q)} \left[ |g(t) - g(q)| + g(q) \right] dt \text{ by definition of absolute value }\\
&\leq \int_{B_r(q)} \left[ \sup_{q' \in B_r(q)}|g(q') - g(q)| + g(q) \right] dt \text{ by definition of } \sup \\
&\leq \int_{q - r}^{q + r} \left[ Lr + g(q) \right] dt \text{ by the Lipschitz constant }\\
&= 2Lr^2 + 2g(q)r.
\end{align*}

Then,
\begin{align}
    \int_0^1 \left|\frac{1}{K}\sum_{k \in KNN_i} F_{Y_k}^{-1}(q) - \mathbb{B}_i^{-1}(q) \right| dq  = \int_0^1 g(q) dq  = \sum_{q \in \Vec{Q}} \int_{B_r(q)} g(t) dt \leq \sum_{q \in \Vec{Q}} [2Lr^2 + 2g(q)r] \tag{Covering Bound} \label{eqn:covering_bound}.
\end{align}

Now, let us return to our original probabilistic statement:
\begin{align*}
    &\mathbb{P}\left( W_1\left(\mathbb{B}_i, \hat{\mathbb{B}}_i \right)  > \varepsilon \right) \\
    &= \mathbb{P}\left( \int_0^1 \left| \mathbb{B}^{-1}_i(q) - \hat{\mathbb{B}}^{-1}_i(q) \right| dq  > \varepsilon \right) \text{ by definition of } W_1 \\
    &\leq \mathbb{P}\left( \sum_{q \in \Vec{Q}} \left[ 2Lr^2 + 2g(q)r \right] > \varepsilon \right) \text{ from Equation~\ref{eqn:covering_bound}} \\
    &\leq \mathbb{P}\left( \sum_{q \in \Vec{Q}} \left[ Lr^2 + g(q)r \right] > \varepsilon/2 \right) \text{ by dividing by 2 on both sides} \\
    &\leq \sum_{q \in \Vec{Q}} \mathbb{P}\left( Lr^2 + g(q)r > r\varepsilon \right) \text{ by Union bound and because there are } \frac{1}{2r} \text{ balls covering } [0,1] \\
    &= \sum_{q \in \Vec{Q}} \mathbb{P}\left( g(q) > \varepsilon - Lr \right) \text{ by subtracting } Lr^2 \text{ and dividing by } r\\
    &= \sum_{q \in \Vec{Q}} \mathbb{P}\left( \left| \mathbb{B}^{-1}_i(q) - \hat{\mathbb{B}}^{-1}_i(q) \right| > \varepsilon - Lr \right) \text{ by definition of } g(q) \\
    &\leq \sum_{q \in \Vec{Q}} 2\exp\left( \frac{-K(\varepsilon - Lr - \delta(\alpha))^2}{2(\zeta_{\max} - \zeta_{\min})^2} \right) \text{ from Equation~\ref{eqn:pointwise_bound}} \\
    &\leq \frac{1}{r} \exp\left( \frac{-K(\varepsilon - Lr - \delta(\alpha))^2}{2(\zeta_{\max} - \zeta_{\min})^2} \right) \text{ because there are } \frac{1}{2r} \text{ balls covering } [0,1].
\end{align*}

\end{proof}

\subsection{Proof of CATE Consistency (Theorem 1)}
\begin{theorem*}[\ref{thm:cate_consistency}]
    Under the same conditions as Lemma \ref{lemma:barycenter_consistency},
    \begin{align*}
        \mathbb{P}\left(\int_0^1 \left|\tau(q | F_{\mathbf{x}_i}) - \hat{\tau}(q | F_{\mathbf{x}_i}) \right| dq > \varepsilon \right)
        \leq 4c\left(\varepsilon/2, \alpha\right).
    \end{align*}
\end{theorem*}

\begin{proof}
    We estimate conditional average treatment effects as a contrast between the quantile functions of the treated and control conditional barycenters. Specifically,
    \begin{align*}
        \hat{\tau}(q) &= \hat{\mathbb{B}}[F_Y|F_{\bm{X}} = F_{\bm{x}_i}, T = 1]^{-1}(q) - \hat{\mathbb{B}}[F_Y|F_{\bm{X}} = F_{\bm{x}_i}, T = 0]^{-1}(q).
    \end{align*}

    We want to show that for all $q \in [0,1], \hat{\tau}(q)$ converges to $\tau(q)$:
    \begin{align*}
        \tau(q) &= \mathbb{B}[F_Y|F_{\bm{X}} = F_{\bm{x}_i}, T = 1]^{-1}(q) - \mathbb{B}[F_Y|F_{\bm{X}} = F_{\bm{x}_i}, T = 0]^{-1}(q).
    \end{align*}
    
    Let $\varepsilon > 0.$ By definitions and rearranging,
    \begin{align*}
        &\mathbb{P}\left(\left\{\int_0^1 \left|\hat{\tau}(q) - \tau(q) \right| dq \right\} > \varepsilon \right) \\
        &= \mathbb{P}\left(\left\{\int_0^1 \left|\begin{pmatrix}
            \hat{\mathbb{B}}[F_Y|F_{\bm{X}} = F_{\bm{x}_i}, T = 1]^{-1}(q) \\
            - \hat{\mathbb{B}}[F_Y|F_{\bm{X}} = F_{\bm{x}_i}, T = 0]^{-1}(q)
        \end{pmatrix} - \begin{pmatrix}
            \mathbb{B}[F_Y|F_{\bm{X}} = F_{\bm{x}_i}, T = 1]^{-1}(q) \\
            - \mathbb{B}[F_Y|F_{\bm{X}} = F_{\bm{x}_i}, T = 0]^{-1}(q)
        \end{pmatrix} \right| dq \right\} > \varepsilon \right) \\
        &= \mathbb{P}\left(\left\{\int_0^1 \left|\begin{pmatrix}
            \hat{\mathbb{B}}[F_Y|F_{\bm{X}} = F_{\bm{x}_i}, T = 1]^{-1}(q) \\
            - \mathbb{B}[F_Y|F_{\bm{X}} = F_{\bm{x}_i}, T = 1]^{-1}(q) 
        \end{pmatrix} - \begin{pmatrix}
            \hat{\mathbb{B}}[F_Y|F_{\bm{X}} = F_{\bm{x}_i}, T = 0]^{-1}(q)\\
            - \mathbb{B}[F_Y|F_{\bm{X}} = F_{\bm{x}_i}, T = 0]^{-1}(q)
        \end{pmatrix} \right| dq \right\} > \varepsilon \right).
    \end{align*}

    By the triangle inequality, we see that
    \begin{align*}
        &\mathbb{P}\left(\left\{\int_0^1 \left|\begin{pmatrix}
            \hat{\mathbb{B}}[F_Y|F_{\bm{X}} = F_{\bm{x}_i}, T = 1]^{-1}(q) \\
            - \mathbb{B}[F_Y|F_{\bm{X}} = F_{\bm{x}_i}, T = 1]^{-1}(q) 
        \end{pmatrix} - \begin{pmatrix}
            \hat{\mathbb{B}}[F_Y|F_{\bm{X}} = F_{\bm{x}_i}, T = 0]^{-1}(q)\\
            - \mathbb{B}[F_Y|F_{\bm{X}} = F_{\bm{x}_i}, T = 0]^{-1}(q)
        \end{pmatrix} \right| dq \right\} > \varepsilon \right) \\
        &\leq \mathbb{P}\left(\left\{\int_0^1 \begin{vmatrix}
            \hat{\mathbb{B}}[F_Y|F_{\bm{X}} = F_{\bm{x}_i}, T = 1]^{-1}(q) \\
            - \mathbb{B}[F_Y|F_{\bm{X}} = F_{\bm{x}_i}, T = 1]^{-1}(q) 
        \end{vmatrix} + \begin{vmatrix}
            \hat{\mathbb{B}}[F_Y|F_{\bm{X}} = F_{\bm{x}_i}, T = 0]^{-1}(q)\\
            - \mathbb{B}[F_Y|F_{\bm{X}} = F_{\bm{x}_i}, T = 0]^{-1}(q)
        \end{vmatrix}  dq \right\} > \varepsilon \right) \\
        &= \mathbb{P}\left(\left\{\int_0^1 \begin{vmatrix}
            \hat{\mathbb{B}}[F_Y|F_{\bm{X}} = F_{\bm{x}_i}, T = 1]^{-1}(q) \\
            - \mathbb{B}[F_Y|F_{\bm{X}} = F_{\bm{x}_i}, T = 1]^{-1}(q) 
        \end{vmatrix}dq + \int_0^1 \begin{vmatrix}
            \hat{\mathbb{B}}[F_Y|F_{\bm{X}} = F_{\bm{x}_i}, T = 0]^{-1}(q)\\
            - \mathbb{B}[F_Y|F_{\bm{X}} = F_{\bm{x}_i}, T = 0]^{-1}(q)
        \end{vmatrix}  dq \right\} > \varepsilon \right).
    \end{align*}
    And by the union bound, we see that
    \begin{align*}
        &\mathbb{P}\left(\left\{\int_0^1 \begin{vmatrix}
            \hat{\mathbb{B}}[F_Y|F_{\bm{X}} = F_{\bm{x}_i}, T = 1]^{-1}(q) \\
            - \mathbb{B}[F_Y|F_{\bm{X}} = F_{\bm{x}_i}, T = 1]^{-1}(q) 
        \end{vmatrix}dq + \int_0^1 \begin{vmatrix}
            \hat{\mathbb{B}}[F_Y|F_{\bm{X}} = F_{\bm{x}_i}, T = 0]^{-1}(q)\\
            - \mathbb{B}[F_Y|F_{\bm{X}} = F_{\bm{x}_i}, T = 0]^{-1}(q)
        \end{vmatrix}  dq \right\} > \varepsilon \right) \\
        &\leq \mathbb{P}\left(\left\{\int_0^1 \begin{vmatrix}
            \hat{\mathbb{B}}[F_Y|F_{\bm{X}} = F_{\bm{x}_i}, T = 1]^{-1}(q) \\
            - \mathbb{B}[F_Y|F_{\bm{X}} = F_{\bm{x}_i}, T = 1]^{-1}(q) 
        \end{vmatrix}dq \right\}> \frac{\varepsilon}{2} \right) + \mathbb{P}\left(\left\{\int_0^1 \begin{vmatrix}
            \hat{\mathbb{B}}[F_Y|F_{\bm{X}} = F_{\bm{x}_i}, T = 0]^{-1}(q)\\
            - \mathbb{B}[F_Y|F_{\bm{X}} = F_{\bm{x}_i}, T = 0]^{-1}(q)
        \end{vmatrix}  dq \right\} > \frac{\varepsilon}{2} \right) \\
        &= \mathbb{P}\left(W_1\begin{pmatrix}
            \hat{\mathbb{B}}[F_Y|F_{\bm{X}} = F_{\bm{x}_i}, T = 1], \\
            \mathbb{B}[F_Y|F_{\bm{X}} = F_{\bm{x}_i}, T = 1]
        \end{pmatrix} > \frac{\varepsilon}{2}\right) + \mathbb{P}\left(W_1 \begin{pmatrix}
            \hat{\mathbb{B}}[F_Y|F_{\bm{X}} = F_{\bm{x}_i}, T = 0], \\
            \mathbb{B}[F_Y|F_{\bm{X}} = F_{\bm{x}_i}, T = 0] 
        \end{pmatrix} > \frac{\varepsilon}{2} \right),
    \end{align*}
    where the last line follows from the definition of 1-Wasserstein in one-dimension.
    
    Let $\alpha$ be the maximum distance between unit $i$ and any of its treated or control K nearest neighbors. From Lemma \ref{lemma:barycenter_consistency}, we know that
    \begin{align*}
        \mathbb{P}\left(W_1\begin{pmatrix}\hat{\mathbb{B}}[F_{Y} | F_{\mathbf{X}} = F_{\mathbf{x}_i}, T = t], 
                \mathbb{B}[F_{Y} | F_{\mathbf{X}} = F_{\mathbf{x}_i}, T = t] \end{pmatrix} > \frac{\varepsilon}{2} \right)
        \leq \frac{1}{r}\exp\left( \frac{-K(\frac{\varepsilon}{2} - Lr - \delta(\alpha))^2}{2(\zeta_{\max} - \zeta_{\min})^2} \right) 
    \end{align*}
    for an $r < \frac{\varepsilon/2 - \delta(\alpha)}{L}.$
    Therefore,
\begin{align*}    
    &\mathbb{P}\left(\left\{\int_0^1 \left|\tau(q | F_{\mathbf{x}_i}) - \hat{\tau}(q | F_{\mathbf{x}_i}) \right| dq \right\}> \varepsilon \right) \\
    &\leq \mathbb{P}\left(W_1\begin{pmatrix}
            \hat{\mathbb{B}}[F_Y|F_{\bm{X}} = F_{\bm{x}_i}, T = 1], \\
            \mathbb{B}[F_Y|F_{\bm{X}} = F_{\bm{x}_i}, T = 1]
        \end{pmatrix} > \frac{\varepsilon}{2} \right) + \mathbb{P}\left(W_1 \begin{pmatrix}
            \hat{\mathbb{B}}[F_Y|F_{\bm{X}} = F_{\bm{x}_i}, T = 0], \\
            \mathbb{B}[F_Y|F_{\bm{X}} = F_{\bm{x}_i}, T = 0] 
        \end{pmatrix} > \frac{\varepsilon}{2} \right) \\
        &\leq \frac{2}{r}\exp\left( \frac{-K(\frac{\varepsilon}{2} - Lr - \delta(\alpha))^2}{2(\zeta_{\max} - \zeta_{\min})^2} \right).
    \end{align*}

\end{proof}

\section{Uncertainty Quantification} \label{sec:uncertainty}

We construct point-wise confidence intervals for the average treatment effect. The motivation and theory follow almost directly from \citet{abadie2011bias}, but we discuss these in the context of our notation and setting with distributional outcomes.

Let $\mu_t\left(q \mid F_{\bm{x}} \right)  = \mathbb{E}\left[F_{Y_i}^{-1}(q) \mid F_{\bm{X}} = F_{\bm{x}}, T = t \right]$ and 
$\sigma^2_t\left(q \mid F_{\bm{x}} \right)  = \mathbb{E}\left[\left(F_{Y_i}^{-1}(q) - \mu_t\left(q \mid F_{\bm{x}} \right) \right)^2 \mid F_{\bm{X}} = F_{\bm{x}}, T = t \right]$. Assume $\muT$ and $\sigmaT$ are smooth: for units $i,j$ with $t_i = t_j$, $\left| \mu_t\left(q \mid F_{\bm{x}_i} \right) - \mu_t\left(q \mid F_{\bm{x}_j} \right)  \right|$  and $\left| \sigma^2_t\left(q \mid F_{\bm{x}_i} \right) - \sigma^2_t\left(q \mid F_{\bm{x}_j} \right)  \right|$ strictly monotonically decrease as $d_{\mathcal{M}}\left( F_{\bm{x}_i}, F_{\bm{x}_j}\right)$ decreases and are 0 when $d_{\mathcal{M}}\left(F_{\bm{x}_i}, F_{\bm{x}_j}\right) = 0$. This assumption places smoothness at the quantile-level rather than at the quantile-\textit{function} level. Also, assume that $\mathbb{E}\left[(F_{Y_i(t)}^{-1}(q))^4 \mid F_{\bm{X}} = F_{\bm{x}_i} \right] \leq C$ for some finite $C$  and that $\sigmaT$ is bounded away from 0. 

Our estimand of interest is the average treatment effect (ATE):
\begin{align*}
    \tau(q) &= \mathbb{E}\left[F_{Y_i(0)}^{-1}(q) - F_{Y_i(0)}^{-1}(q) \right].
\end{align*}

Let $\hatmuT$ represent a consistent estimator of $\muT.$ Also, let $M_K(i) = \sum_{l = 1}^{\left|\mathcal{S}_{est} \right|} \mathbf{1}\left[ i \in KNN_{d_{\mathcal{M}}}\left(F_{\bm{x}_l}, \mathcal{S}_{est}^{(1-t_l)}\right) \right]$ represent the number of times unit $i$ is in \textit{another} unit's set of $K$ nearest neighbors of the opposite treatment status. Rewriting our estimate for the ATE:
\begin{align*}
    \hat{\tau}(q) &= \frac{1}{\left|\mathcal{S}_{est} \right|} \sum_{i \in \mathcal{S}_{est}} \left(2T_i - 1 \right) \cdot \left(1 + \frac{M_K(i)}{K} \right) \cdot F_{Y_i}^{-1}(q) \\
    &= \frac{1}{\left|\mathcal{S}_{est} \right|}\sum_{i \in \mathcal{S}_{est}} \hat{F}_{Y_i(1)}^{-1}(q) - \hat{F}_{Y_i(0)}^{-1}(q),
\end{align*}
where
\begin{align*}
    \hat{F}_{Y_i(t)}^{-1}(q) &= \begin{cases} 
        F_{Y_i}^{-1}(q) &\text{if } t = T_i \\
        \frac{1}{K}\sum_{j \in KNN_{d_{\mathcal{M}}}\left(F_{\bm{x}_i}, \mathcal{S}_{est}^{(1 - t)} \right)} F_{Y_j}^{-1}(q) &\text{if } t \neq T_i.
    \end{cases}
\end{align*}
However, $\hat{\tau}(q)$ is biased; \citet{abadie2011bias} provide a closed form solution for the bias term and an estimator:
\begin{align*}
    B(q) &= \frac{1}{\left|\mathcal{S}_{est} \right|} \sum_{i \in \mathcal{S}_{est}} \frac{2T_i - 1}{K} \sum_{j \in KNN_{d_{\mathcal{M}}}\left(F_{\bm{x}_i}, \mathcal{S}_{est}^{1-T_i} \right)} \mu_{1-T_i}\left(q \mid F_{\bm{x}_i} \right) - \mu_{1-T_i}\left(q \mid F_{\bm{x}_j} \right).
\end{align*}
We then have an unbiased estimate of the ATE, $\tau(q) = \mathbb{E}\left[\hat{\tau}(q) - B(q) \right].$ However, we do not know the true conditional mean function $\muT,$ so we must estimate them. Assume we have consistent estimators of the conditional means of the outcomes. Then, we can consistently estimate B(q):
\begin{align*}
    \hat{B}(q) &= \frac{1}{\left|\mathcal{S}_{est} \right|} \sum_{i \in \mathcal{S}_{est}} \frac{2T_i - 1}{K} \sum_{j \in KNN_{d_{\mathcal{M}}}\left(F_{\bm{x}_i}, \mathcal{S}_{est}^{1-T_i} \right)} \hat{\mu}_{1-T_i}\left(q \mid F_{\bm{x}_i} \right) - \hat{\mu}_{1-T_i}\left(q \mid F_{\bm{x}_j} \right)
\end{align*}
We can then construct our bias-corrected estimator $\hat{\tau}^{bcm}(q) = \hat{\tau}(q) - \hat{B}(q).$

We now construct the variance of our bias corrected estimator. Let $\left\{F_{\bm{X}_i}, T_i\right\}_{i \in \mathcal{S}_{est}}$ represent the covariates and treatment statuses observed in the estimation set. And let
\begin{align*}
    V^{E}(q) &= \frac{1}{\left|\mathcal{S}_{est} \right|}\mathbb{V}\left[\hat{\tau}(q) \mid  \left\{F_{\bm{X}_i}, T_i\right\}_{i \in \mathcal{S}_{est}} \right] \\
    &= \frac{1}{N}\sum_{i = 1}^N \left(1 + \frac{M_K(i)}{K} \right)^2 \sigma^2_{T_i}\left(q \mid F_{\bm{x}_i} \right), \\
    V^{\tau(q|F_{\bm{x}_i})}(q) &= \mathbb{E}\left[ \left( \left[\muTreated - \muControl \right] - \tau(q) \right)^2 \right].
\end{align*}
Then,
\begin{align*}
    \sqrt{\left| \mathcal{S}_{est} \right|}\left(V^{E}(q) + V^{\tau(q | F_{\bm{x}_i})}(q) \right)^{-1/2}\left(\hat{\tau}^{bcm}(q) - \tau(q) \right) \overset{d}{\to} \mathcal{N}(0,1).
\end{align*}
By estimating $V^{E}(q) + V^{\tau(q | F_{\bm{x}_i})}(q),$ we could construct $1-\alpha \%$ confidence intervals. To do this, we must first find a way to estimate $\sigma^2_{T_i}\left(q \mid F_{\bm{x}_i} \right),$  the conditional variance of the outcome. Let $\ell_j(i)$ represent the $j^{th}$ nearest neighbor of $i$ with the same treatment status in the estimation set. For some fixed $J,$ we can estimate $\sigma^2_{T_i}\left(q \mid F_{\bm{x}_i} \right)$ as
\begin{align*}
    \hat{\sigma}^2_{T_i}\left(q \mid F_{\bm{x}_i} \right) &= \frac{J}{J + 1}\left(F_{Y_i}^{-1}(q) - \frac{1}{J} \sum_{j = 1}^{J} F_{Y_{\ell_j(i)}}^{-1}(q) \right)^2.
\end{align*}
Theorem 7 of \citet{abadie2006large} then shows that we can estimate valid confidence intervals using the following consistent estimator of $V(q) = V^{E}(q) + V^{\tau(q | F_{\bm{x}_i}})(q):$
\begin{align}
\label{eqn:abadie_imbens_variance}
    \hat{V}(q) = &\frac{1}{\left|\mathcal{S}_{est}\right|}\sum_{i \in \mathcal{S}_{est}} \left|\hat{F}_{Y_i(1)}^{-1}(q) - \hat{F}_{Y_i(0)}^{-1}(q) - \hat{\tau}(q) \right|^2 \notag \\
    + &\frac{1}{\left|\mathcal{S}_{est}\right|}\sum_{i \in \mathcal{S}_{est}} \left[\left(\frac{M_K(i)}{K} \right)^2 + \left(\frac{2K - 1}{K} \right)\left(\frac{M_K(i)}{K} \right) \right]\hat{\sigma}^2_{T_i}\left(q \mid F_{\bm{x}_i} \right).
\end{align}

\subsection{Simulation Study}
We use the data generating processes with scalar covariates in Table \ref{tab:dgps} to evaluate the coverage of our uncertainty quantification strategy. For each of 100 Monte Carlo iterations, we first calculate the true average treatment effect and use the variance estimator in Equation~\ref{eqn:abadie_imbens_variance} to construct 95\% \textit{pointwise} confidence intervals. We then evaluate the nominal pointwise coverage.

\begin{table}[ht]
    \centering
    \begin{tabular}{c|c}
         DGP & Coverage \\
         \hline
        Variance  & 0.947  \\
         Linear   & 0.963 \\
         Complex  & 0.989  
    \end{tabular}
    \caption{The estimated coverage of 95\% confidence intervals computed using the variance from Equation~\ref{eqn:abadie_imbens_variance}.}
    \label{tab:coverage}
\end{table}

\section{Real Data Analysis: Exploratory Data Analysis}
In this section, we further explore the data from \citet{juvenile2010effectiveness, juvenile2008continuous}. Specifically, we first show how we clean the data; and because the trial randomized treatment, we compare ADD MALTS' estimated ATE to the difference-in-means-estimated ATE to validate that ADD MALTS can also recover an accurate ATE estimate.

\paragraph{Data Cleaning and Processing}
The data analyses were published in two separate studies: \citet{juvenile2008continuous} studies patients for whom glycated blood hemoglobin levels (HbA1c) at baseline was greater than 8\% and \citet{juvenile2010effectiveness} studies patients for whom glycated blood hemoglobin levels (HbA1c) at baseline was $\leq 8\%.$ Both studies excluded patients who did not complete the full 26 weeks of randomization; we also excluded patients that did not have CGM readings more than 26 weeks worth of readings after the last recording in their baseline data. We constructed each quantile function using 900 quantiles; to exclude outlier glucose readings, we only evaluate treatment effects between the 2.5 and 97.5 percentiles.

\begin{figure}
    \centering
    \includegraphics[width=0.7\textwidth]{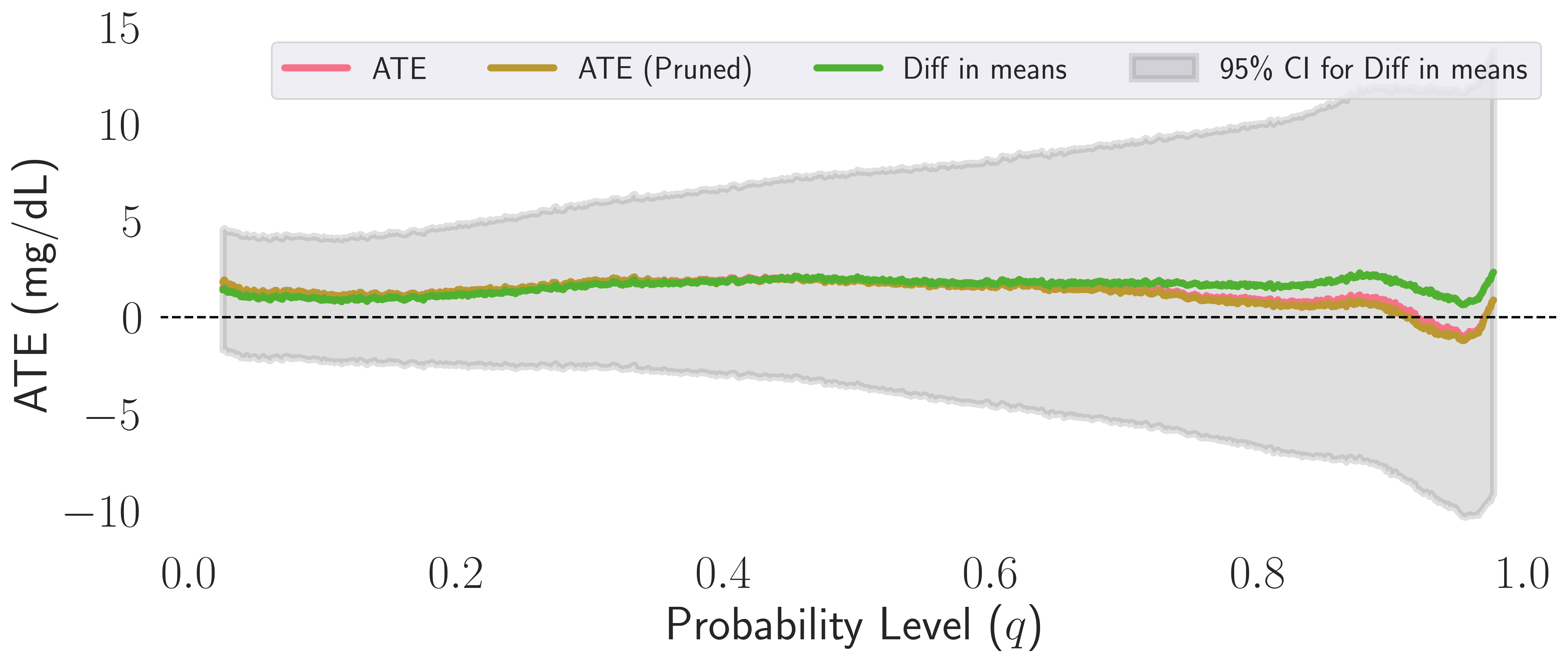}
    \caption{We compare the difference-in-means-estimated ATE (green) to the ADD MALTS' estimated ATE before and after pruning bad matched groups (pink, gold respectively). The grey interval represents 95\% confidence interval for the difference-in-means-estimated ATE.}
    \label{fig:ate_ttest}
\end{figure}

\paragraph{ADD MALTS vs Difference of Means}
Because the data we analyze is from a randomized experiment, the difference in mean quantile functions will be an unbiased estimate of average treatment effects. We compare ADD MALTS' estimated ATE to the difference-in-means-estimated ATE (DIME ATE). As seen in Figure \ref{fig:ate_ttest}, the two estimates are not significantly different from one another (the ADD MALTS ATE is within the 95\% confidence interval for the DIME ATE). Also, there is a very marginal difference between the point-estimates: the DIME ATE also ranges between -2 and 2, and the difference between the DIME and ADD MALTS' ATEs range between -1 and 1. The fact that ADD MALTS' estimated ATE is so close to the DIME ATE validates ADD MALTS' ability to estimate average treatment effects in real-world settings.

\section{Additional Synthetic Experiments}
In this section, we provide more details on the CATE estimation experiments and the positivity violation experiment. We also demonstrate how ADD MALTS' CATE estimation performance is insensitive to the number of nearest neighbors. 

\subsection{CATE Estimation Experimental Details} \label{sec:cate_exp_details}
In this section, we provide more details on CATE estimation experiments in the main paper.

\begin{table*}[ht]
    \centering
    \begin{tabular}{c|c|c}
        \textbf{Name} & $\mathbf{\mu}$ & $\mathbf{\sigma}$  \\
        \hline
        Variance & $10 + x_{i,0} + 2x_{i,1} + \varepsilon_i$ & $\left|10 + x_{i,0} + 2x_{i,1} + 10T_i + \varepsilon_i \right|$ \\
        \hline
        Linear & $10 + x_{i,0} + 2x_{i,1} + 10T_i + \varepsilon_i$ & 1 \\
        \hline
        Complex & $\begin{matrix}
        10\sin\left(\pi x_{i,0}x_{i,1} \right) + 20(x_{i,2} - 0.5)^2 + 10x_{i,3} + 5x_{i,4} \\
        + T_i\left\{7 + x_{i,2}\cos\left(\pi x_{i,0}x_{i,1}\right) \right\} + \varepsilon_i
                    \end{matrix} $ & $\left|10 + x_{i,0} + 2x_{i,1} + \varepsilon_i\right|$ \\
        \hline
        Dist Cov & $\begin{matrix}
        10\sin\left(\pi \left[\int_0^1F_{x_{i,0}}^{-1}(q) dq \right]x_{i,1} \right) + 20(x_{i,2} - 0.5)^2 + 10x_{i,3} + 5x_{i,4} \\
        + T_i\left\{7 + x_{i,2}\cos\left(\pi \left[\int_0^1F_{x_{i,0}}^{-1}(q) dq \right]x_{i,1}\right) \right\} + \varepsilon_i
                    \end{matrix} $ & 1
    \end{tabular}
    \caption{In our simulations, the distributional outcomes are truncated normal distributions (truncated at $\pm 3$ standard deviations from the mean). The table describes the functions used to generate the mean and variance of each outcome.  In the ``Linear,'' ``Variance,'' and ``Complex'' DGPs, we respectively sample 2, 2, and 6 \textit{scalar} covariates independently and identically from \textit{Uniform}$[-1,1].$ The last simulation (Dist Cov) has a distributional covariate $F_{x_{i,0}}$, which we use by taking the integral of its cumulative distribution function as a covariate, and 9 scalar covariates.}
    \label{tab:dgps}
\end{table*}

We consider four data generative processes (DGPs). In each DGP, we generate our distributional outcomes as truncated normal distributions (truncated at $\pm 3$ standard deviations from the mean). Table \ref{tab:dgps} describes the functions used to generate the means and variances of the truncated normal outcomes for each DGP. In the ``Linear,'' ``Variance,'' and ``Complex'' DGPs, we respectively sample 2, 2, and 6 \textit{scalar} covariates independently and identically from \textit{Uniform}$[-1,1].$ The propensity score models are linear: $\mathbb{P}(T_i = 1 | F_{\bm{x}_i}) = \text{expit}(x_{i,0} + x_{i,1}).$

The ``Dist Cov'' simulation instead has nine scalar covariates and a distributional covariate $F_{x_{i,0}}$. $F_{x_{i,0}}$ is uniformly sampled to be any uniform distribution between $[-1,0]$ and $[-1,1].$ We use the integral of the quantile function as a feature to generate outcomes. Because baseline methods can only handle scalar covariates, we instead provide them the preprocessed covariate, the area under the quantile function. Here, the propensity score model is also linear but depends on this processed distribution: $\mathbb{P}(T_i = 1|F_{\bm{x}_i}) = \text{expit}\left(\left[\int_0^1 F_{x_{i,0}}^{-1}(q) dq\right] + x_{i,2} \right).$

For each DGP, we consider 1500 units, using $67\%$ in the training set and $33\%$ to estimate CATEs. Also, each DGP has at least five irrelevant covariates. We repeat each experiment 100 times and evaluate CATE estimation performance using the $\text{Integrated Relative Error (\%)} = 100 \cdot \int_0^1 \left| \frac{\hat{\tau}(q) - \tau(q)}{\tau(q)} \right|dq$.

We compare ADD MALTS to the following baselines: \textbf{Lin PSM} and \textbf{RF PSM} represent propensity score matching fit with linear and random forest models, respectively. We use a cross-validated $\ell_1$-regularized logistic regression to train the linear propensity score model. We use the default settings in \texttt{sklearn}'s random forest implementation to train the random forest propensity score model. \textbf{FT} and \textbf{FRF} represent decision tree and random forest methods for functional outcomes \citep{qiu2022random}; we use a depth bound of 5 to train the decision tree and 100 trees of depth bound 20 to train our forest\footnote{The choice of 100 trees comes from the default setting in \texttt{sklearn}'s implementation of random forests. While \texttt{sklearn} has no depth bound, the functional outcome tree code we wrote ran into memory issues when the depth bound was greater than 20}. \textbf{LR} represents outcome regression fit at each quantile with a linear regression \citep{lin2023causal}; \textbf{LR + Lin PS} and \textbf{LR + RF PS} represent augmented inverse propensity weighting methods combining the linear outcome regression with linear and random forest propensity score models, respectively. Here, we use an unregularized linear regression for the outcome model, as in \citet{lin2023causal}, and use the same propensity score model configurations as in the propensity score matching methods. Table~\ref{tab:code_dependencies} details the package dependencies of each method that we used for implementation.

\subsubsection{Timing Results} \label{sec:timing}
\begin{figure}[ht]
    \centering
    \includegraphics[width=0.4\textwidth]{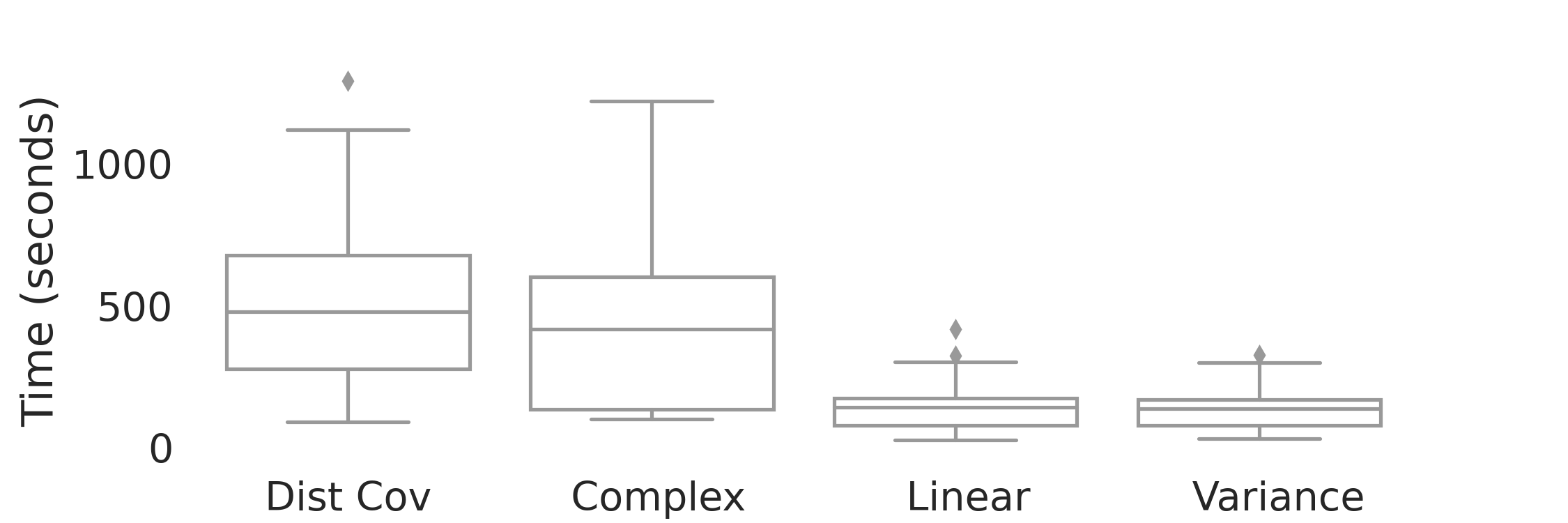}
    \caption{Each boxplot displays the time in seconds (y-axis) it took for ADD MALTS to run, from fitting the distance metric to estimating CATEs, for each DGP in Table~\ref{tab:dgps} across 100 Monte Carlo iterations.}
    \label{fig:timing}
\end{figure}
In this section, we evaluation ADD MALTS' running time. ADD MALTS had median run-time across all Monte Carlo iterations less than 10 minutes, from fitting the distance metric to estimating CATEs (see Figure \ref{fig:timing}). The Linear and Variance DGPs had the lowest median runtime, less than three minutes per iteration.

\subsection{ADD MALTS' CATE Estimation Performance on Another DGP}
\begin{algorithm}[ht]
\caption{Mixture-beta outcomes DGP} \label{alg:mix_beta_dgp}
\begin{algorithmic}
\For{$i = 1, \ldots, 1000$}{
    \State $X_{i,1}, \ldots, X_{i,10} \sim \text{Unif}[-1,1]$ \Comment{Generate scalar covariates}
    \State $T_i \sim \text{Bern}\left(\text{expit}\left( X_{i,1} + X_{i,2}\right) \right)$ \Comment{Assign treatment} 
    
    \State $\alpha(t_i) = 5 + 10\sin(\pi X_{i,1}X_{i,2})^2 + 20(X_{i,3} - 0.5)^2 + 10X_{i,4} + 5X_{i,5} + t_i10X_{i,3}\cos(\pi X_{i,1}X_{i,2})^2 + \varepsilon_{Y_i}, \varepsilon_{Y_i} \sim \mathcal{N}(0,1)$ \Comment{Generate parameter that will control Beta distribution's shape}
    \For{$j = 1, \ldots, 1001$}{
        \State $Z_{i,j} \sim \text{Bern}(1/4)$ \Comment{Sample which mixture of the outcome this observation will come from}
        \State $Y_{i,j}(0) \sim \text{Beta}(2\alpha(0), 8\alpha(0))^{Z_{i,j}}\text{Beta}(8\alpha(0), 2\alpha(0))^{1-Z_{i,j}}$ \Comment{Generate sample from the control potential outcome}
        \State $Y_{i,j}(1) \sim \text{Beta}(2\alpha(1), 8\alpha(1))^{1 - Z_{i,j}}\text{Beta}(8\alpha(1), 2\alpha(1))^{Z_{i,j}}$ \Comment{Generate sample from the treated potential outcome}
        \EndFor
        }
    \State $F_{Y_i(0)}^{-1}(q) = \min\{y : \frac{1}{1001} \sum_{j = 1}^{1001} \mathbf{1}[Y_{i,j}(0) \leq y] \geq q\}$ \Comment{Generate the control potential outcome's quantile function}
    \State $F_{Y_i(1)}^{-1}(q) = \min\{y : \frac{1}{1001} \sum_{j = 1}^{1001} \mathbf{1}[Y_{i,j}(1) \leq y] \geq q\}$ \Comment{Generate the treated potential outcome's quantile function}
    \State $\tau(q|F_{\bm{x}_i}) = F_{Y_i(1)}^{-1}(q) - F_{Y_i(0)}^{-1}(q)$ \Comment{Calculate the true CATE}
    \State $F_{Y_i}^{-1}(q) = T_iF_{Y_i(1)}^{-1}(q) + (1 - T_i)F_{Y_i(0)}^{-1}(q)$ \Comment{Calculate the observed outcome}
\EndFor}

\Return $\mathcal{S}_n = \left\{ (X_{i,1}, \ldots, X_{i,10}), T_i, F_{Y_i}^{-1} \right\}_{i = 1}^{n}$  \Comment{Return observations}
\end{algorithmic} 
\end{algorithm}
\begin{figure}[ht]
     \centering
     \begin{subfigure}[b]{0.45\textwidth}
         \centering
         \includegraphics[width=\textwidth]{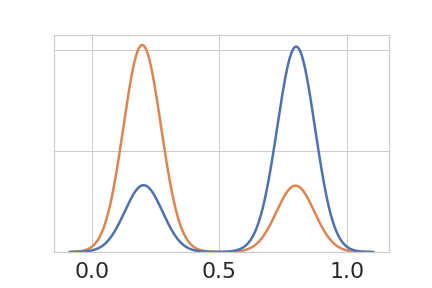}
         \caption{}
     \end{subfigure}
     \hfill
     \begin{subfigure}[b]{0.45\textwidth}
         \centering
         \includegraphics[width=\textwidth]{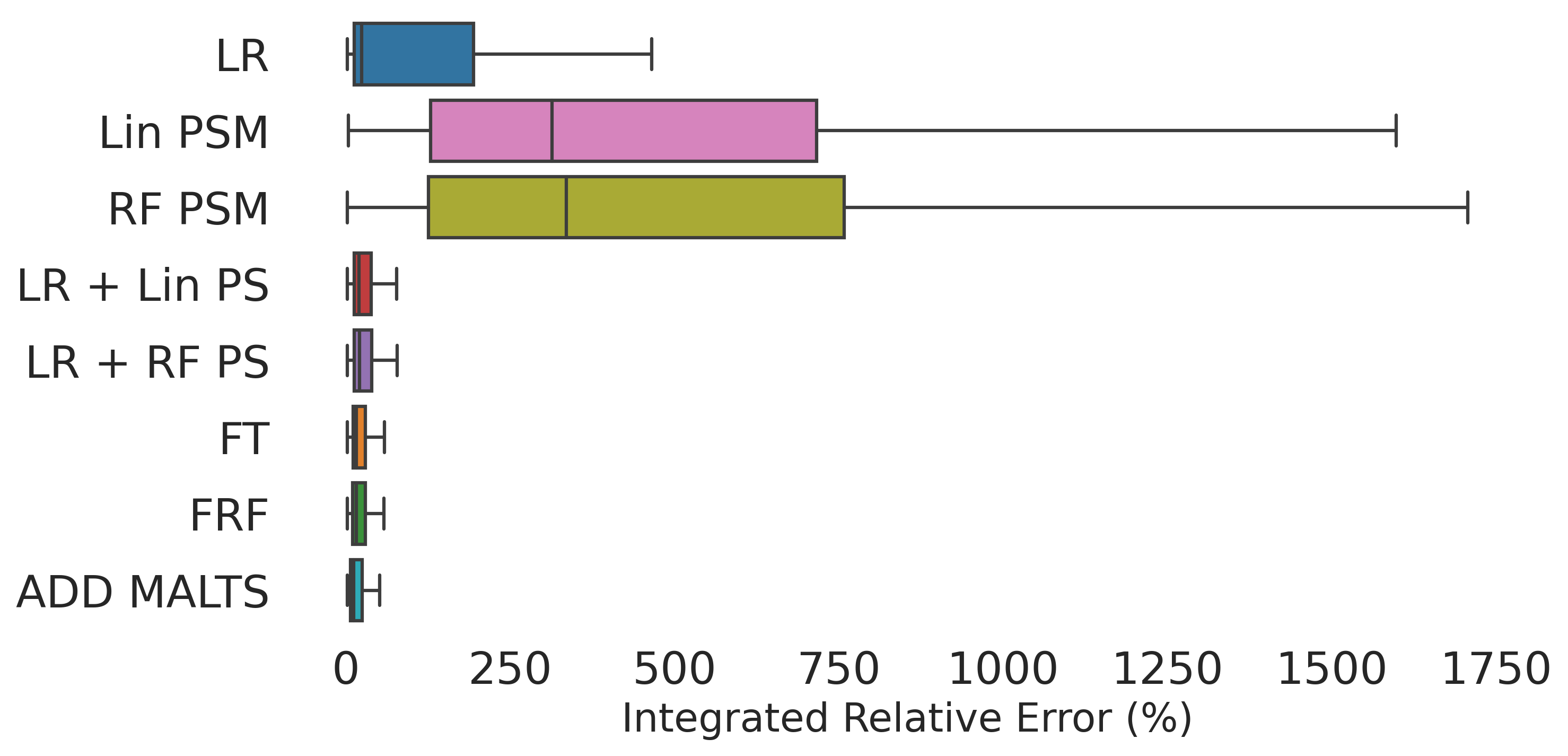}
         \caption{}
     \end{subfigure}
        \caption{Subplot (a) displays the densities of the treated (blue) and control (orange) potential outcomes generated from this data generating process. Subplot (b) displays the integrated relative error (y-axis) for the various baselines and ADD MALTS for the DGP described in Algorithm~\ref{alg:mix_beta_dgp}.}
        \label{fig:beta_mixture_details}
\end{figure}

In this section, we also evaluate ADD MALTS' CATE estimation performance when the outcome distribution is not a truncated normal distribution. In this experiment, we generate our outcomes as a mixture of two-beta distributions (as seen in Algorithm~\ref{alg:mix_beta_dgp}).

The DGP has 10 scalar covariates. We then assign treatment using a simple, linear propensity score model using two of the covariates. Our outcome is a mixture of two Beta distributions whose parameters are controlled by the term $\alpha(t_i)$; $\alpha(t_i)$ is a combination of complex quadratic and trignometric terms. When $t_i = 1,$ the mixture proportions flip so that more mass is concentrated in the upper end of the distribution's support; additionally, when $t_i = 1$, the Beta distribution's parameter increases by another complex interaction of trignometric terms, causing the variance of each mixture component to shrink. We construct each unit's observed quantile function using the outcome observations associated with unit $i$'s treatment status: $y_{i,1}(t_i), \ldots, y_{i,1001}(t_i)$. In our DGP, we have 5 relevant covariates and 5 irrelevant covariates. We also use a 60/40 train/estimation split to estimate conditional average treatment effects. As seen in Figure~\ref{fig:beta_mixture_details}, ADD MALTS does at least as well as the other methods in this complex DGP. 

\subsection{Positivity Violation Experimental Details} \label{sec:positivity_exp_details}
In this section, we provide more details on the Positivity Violations experiment in Section \ref{sec:overlapDiscovery}. While we describe the DGP we considered in the main paper, we offer more details on the implementation of the baselines and ADD MALTS in this section. We have two baseline methods: a linear propensity score model fit using cross-validation (using the default LASSO logistic regression cross-validation parameters in \texttt{sklearn}) and a random forest propensity score model fit using cross-validation (cross-validating over the number of trees: 20, 50, 100, 200). We use \texttt{sklearn}'s cross-validation implementations of both these techniques to find the best parameters.

\begin{figure}[ht]
    \centering
    \includegraphics[width=0.7\textwidth]{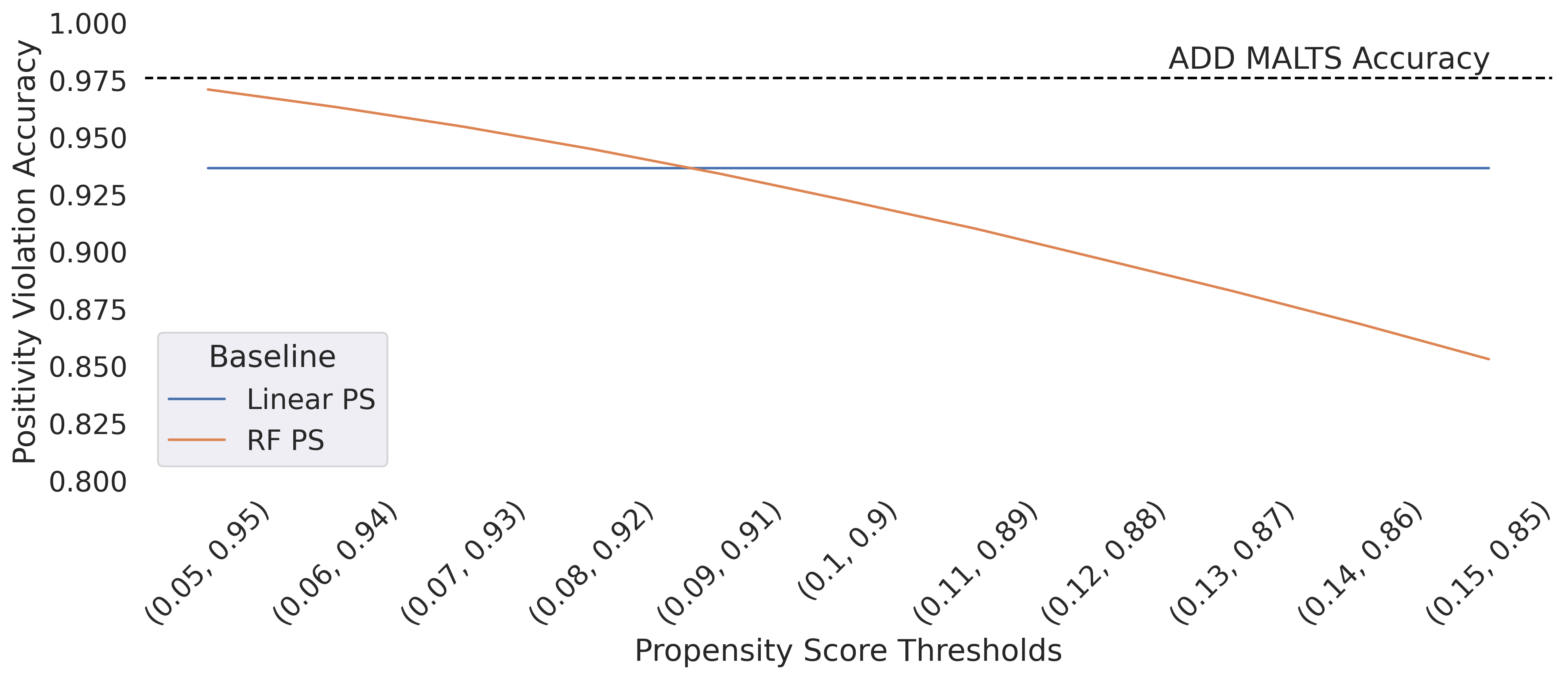}
    \caption{The x-axis represents the thresholds used for flagging regions of the covariate space that may suffer from a positivity violation using the propensity score methods. The y-axis represents each method's accuracy in flagging units that are in positivity violation regions. The orange line represents the accuracy for the random forest estimated propensity score while the blue line represents the accuracy for the propensity score estimated with a linear model. The black, dotted line represents the accuracy when using ADD MALTS' diameter to flag units.}
    \label{fig:overlap_prop_acc}
\end{figure}

\paragraph{Propensity score flagging methods and their relationship to the propensity score thresholds} We evaluate how sensitive RF PS and Lin PS are to changes in the propensity score thresholds when flagging units as being in positivity violation regions. As evidenced in Figure~\ref{fig:overlap_prop_acc}, ADD MALTS outperforms the other methods with various propensity score thresholds. Furthermore, the qualitative inspection of nearest neighbor sets using ADD MALTS offers a layer of fidelity unachievable by the propensity score methods that produce uninterpretable matched groups (see \citet{parikh2022malts} for more details on this comparison).

\subsection{Insensitivity to the Number of Nearest Neighbors}

\begin{figure}[ht!]
    \centering
    \includegraphics[width=0.7\textwidth]{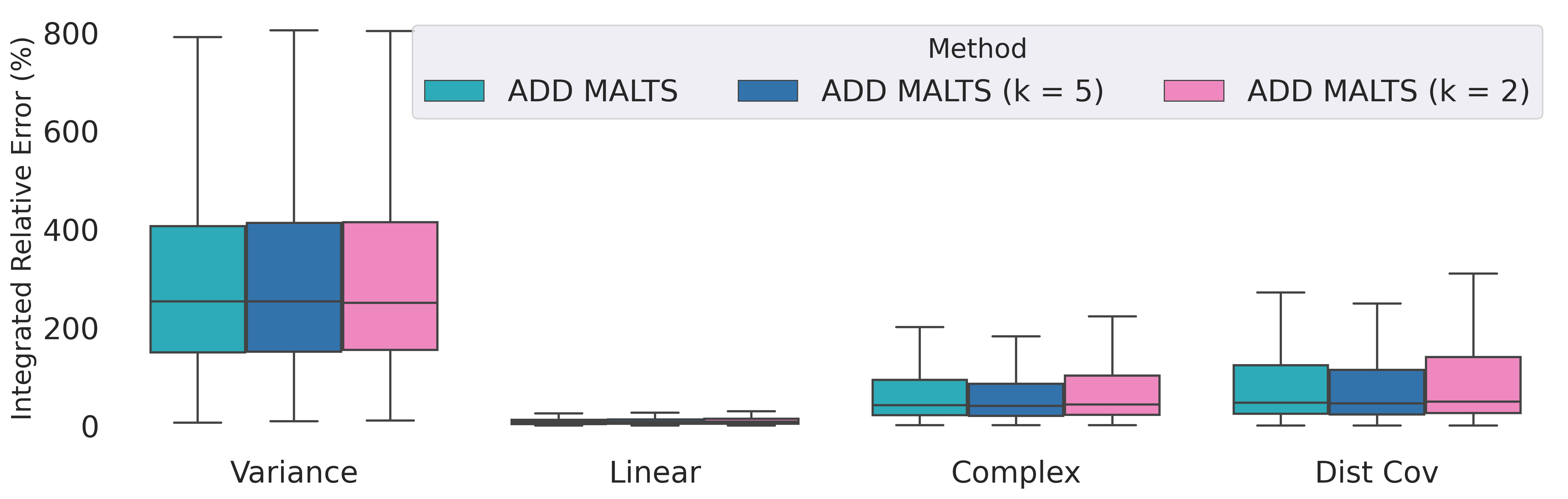}
    \caption{The figure displays the integrated relative error (\%) on the y-axis of each DGP (x-axis) for ADD MALTS estimators with different numbers of nearest neighbors (10 in turquoise, 5 in navy blue, 2 in pink).}
    \label{fig:k_comparison}
\end{figure}

In this section, we demonstrate that ADD MALTS' CATE estimation performance is not affected by the choice of the number of nearest neighbors. Figure~\ref{fig:k_comparison} displays the integrated relative error (\%) on the y-axis of each DGP (x-axis) for ADD MALTS estimators with different numbers of nearest neighbors (10, 5, 2). All box-plots overlap, with the largest difference being a deviation of about 50\% IRE between $k = 5$ and $k = 2$ in the ``Dist Cov'' simulation. A 50\% difference in IRE is marginal compared to the range of values seen in Figure~\ref{fig:cate_estimation_collated} of 0-1500\%. 

\subsection{Computational Resources} \label{sec:comp_resources}
All experiments for this work were performed on an academic institution’s cluster computer. We used up to 40 machines in parallel, selected from the specifications below:
\begin{itemize}
    \item 2 Dell R610’s with 2 E5540 XeonProcessors (16cores)
    \item 10 Dell R730’s with 2 Intel Xeon E5-2640 Processors (40 cores)
    \item 10 Dell R610’s with 2 E5640 Xeon Processors (16 cores)
    \item 10 Dell R620’s with 2 Xeon(R) CPU E5-2695 v2’s (48 cores)
    \item 8 Dell R610’s with 2 E5540 Xeon Processors (16cores)
\end{itemize}
We did not use GPU acceleration for this work.

\begin{table}[ht]
    \centering
    \begin{tabular}{c|c}
        Method & Code Dependencies \\
        \hline
        Lin PSM & \citet{pedregose2011scikit, harris2020array} \\
        RF PSM  & \citet{pedregose2011scikit, harris2020array} \\
        FT      & \citet{pedregose2011scikit, harris2020array, reback2020pandas} \\
        FRF     & \citet{pedregose2011scikit, harris2020array, reback2020pandas} \\
        LR      & \citet{pedregose2011scikit, harris2020array} \\
        LR + Lin PS & \citet{pedregose2011scikit, harris2020array} \\
        RF + Lin PS & \citet{pedregose2011scikit, harris2020array} \\
        ADD MALTS & \citet{harris2020array, 2020SciPy-NMeth, reback2020pandas}
    \end{tabular}
    \caption{A table describing which libraries each baseline in the CATE estimation experiment depended on.}
    \label{tab:code_dependencies}
\end{table}

\newpage

\end{appendices}
\newpage

\begin{@fileswfalse}
\bibliography{biblio}
\end{@fileswfalse}

\end{document}